%% file: full-paper-template.tex
\documentclass[pmlr]{jmlr}

\RequirePackage{graphicx}
 \usepackage{booktabs}
 
 \usepackage{appendix}
 \usepackage{hyperref}
 
\usepackage{tablefootnote}
\usepackage{multirow}
\usepackage{multicol}
\usepackage{array}
\usepackage{float} 
 \usepackage{tikz}
\usepackage{longtable}
\makeatletter
\def\set@curr@file#1{\def\@curr@file{#1}} 
\makeatother
\usepackage[load-configurations=version-1]{siunitx} 

\theorembodyfont{\upshape}
\theoremheaderfont{\scshape}
\theorempostheader{:}
\theoremsep{\newline}

\jmlrvolume{252}
\jmlryear{2024}
\jmlrworkshop{Machine Learning for Healthcare}

\makeatletter
\RenewDocumentCommand{\appendix}{} %
  {
    \gdef\theHchapter{\Hy@AlphNoErr{chapter}} 
    \gdef\theHsection{\Hy@AlphNoErr{section}} 
    \xdef\Hy@chapapp{\Hy@appendixstring} 
    \def\Hy@chapterstring{section} 
    \HyOrg@appendix 
  }
\renewcommand{\thefootnote}{\fnsymbol{footnote}}

\makeatother

\renewcommand{\thefootnote}{\fnsymbol{footnote}}
\newcommand{\dgr}{\textsuperscript{\textdagger}}
\newcommand{\multicol}[3]{
    \multicolumn{1}{r@{\hspace*{0.75\tabcolsep}}}{#1} &
    \multicolumn{1}{l@{\hspace*{0.75\tabcolsep}}}{#2} &
    {#3}
}
\newcommand{\dualcol}[2]{
    \multicolumn{1}{r@{\hspace*{\tabcolsep}\makebox[0pt]{/}}}{#1} &
    {#2}
}
\newcommand{\behrtdata}{$\mathrm{BEHRT}_{+\mathrm{D}}\,$}
\newcommand{\behrtbest}{CORE-BEHRT}

\title[CORE-BEHRT]{CORE-BEHRT: A Carefully Optimized and Rigorously Evaluated BEHRT}
\author{
\Name{Mikkel Odgaard}*\textsuperscript{1}
\Email{miod@di.ku.dk}\\
\Name{Kiril Klein}*\textsuperscript{1}
\Email{kikl@di.ku.dk}\\
\Name{Sanne Møller Thysen}\textsuperscript{2}
\Email{sanne.marie.thysen.01@regionh.dk}\\
\Name{Espen Jimenez-Solem}\textsuperscript{2,3,4}
\Email{espen.jimenez.solem@regionh.dk}\\
\Name{Martin Sillesen}\textsuperscript{5}
\Email{martin.hylleholt.sillesen@regionh.dk}\\
\Name{Mads Nielsen}\textsuperscript{1}
\Email{madsn@di.ku.dk}\\
\normalfont{\small{\textit{*Equal contribution, order determined by coin toss}}} \\
\normalfont{\small{\textit{\textsuperscript{1}Department of Computer Science, University of Copenhagen}}} \\
\normalfont{\small{\textit{\textsuperscript{2}Department of Clinical Pharmacology, Copenhagen University Hospital - Bispebjerg and Frederiksberg}}} \\
\normalfont{\small{\textit{\textsuperscript{3}Copenhagen Phase IV Unit (Phase4CPH), Department of Clinical Pharmacology, Center for Clinical Research and Prevention, Copenhagen University Hospital - Bispebjerg and Frederiksberg}}} \\
\normalfont{\small{\textit{\textsuperscript{4}Department of Clinical Medicine, University of Copenhagen}}} \\
\normalfont{\small{\textit{\textsuperscript{5}Department of Organ Surgery and Transplantation, Copenhagen University Hospital - Rigshospitalet}}}
}

\begin{document}

\maketitle

\input{sections/abstract}
\footnotetext[0]{\hspace{-1em}The code is available on https://github.com/mikkelfo/CORE-BEHRT}

\input{sections/introduction}
\input{sections/related_works}
\input{sections/methods}
\input{sections/cohort}
\input{sections/results}

\input{sections/discussion}


\bibliography{biblo}
\newpage
\appendix
\input{sections/appendix}
\end{document}

%% file: sections/abstract.tex
\begin{abstract}
The widespread adoption of Electronic Health Records (EHR) has significantly increased the amount of available healthcare data. This has allowed models inspired by Natural Language Processing (NLP) and Computer Vision, which scale exceptionally well, to be used in EHR research. 
Particularly, BERT-based models have surged in popularity following the release of BEHRT and Med-BERT. Subsequent models have largely built on these foundations despite the fundamental design choices of these pioneering models remaining underexplored. 
Through incremental optimization, we study BERT-based EHR modeling and isolate the sources of improvement for key design choices, giving us insights into the effect of data representation, individual technical components, and training procedure. Evaluating this across a set of generic tasks (death, pain treatment, and general infection), we showed that improving data representation can increase the average downstream performance from 0.785 to 0.797 AUROC ($p<10^{-7}$), primarily when including medication and timestamps. Improving the architecture and training protocol on top of this increased average downstream performance to 0.801 AUROC ($p<10^{-7}$).
We then demonstrated the consistency of our optimization through a rigorous evaluation across 25 diverse clinical prediction tasks. We observed significant performance increases in 17 out of 25 tasks and improvements in 24 tasks, highlighting the generalizability of our results.
Our findings provide a strong foundation for future work and aim to increase the trustworthiness of BERT-based EHR models.
\end{abstract}

%% file: sections/introduction.tex
\section{Introduction}
In recent years, leveraging Electronic Health Records (EHR) to predict patient outcomes has gained significant interest. Initially, progress was gradual due to a select number of available machine learning methods, such as logistic regression and random forest \citep{fine1997logisticregression, khalilia2011predicting_rf}. 
Although models like XGBoost \citep{chen2016xgboost} show impressive performance on tabular data, the field has focused largely on adopting deep learning models from natural language processing (NLP) and computer vision. These models, coupled with the drastic increase in available EHR data \citep{slawomirski2023progress}, allow sequential processing of the data, taking into account its temporal aspect. \\ \indent
This has been exemplified by the use of recurrent \citep{choi2015doctorai, choi2016usingrn, choi2016retain} and convolutional \citep{nguyen2016deepr} neural networks on EHR data. The introduction of the transformer architecture \citep{vaswani2017attention}, known for its versatility across different data modalities and exceptional scaling, significantly impacted EHR analysis. Specifically, building on the seminal work of BERT (Bidirectional Encoder Representations from Transformers) \citep{devlin2018bert}, the field has seen the emergence of transformer-based models tailored to EHR data \citep{li2020behrt, rasmy2021medbert}. These early models inspired a wave of subsequent research introducing a variety of modifications touching on input data representation, model architecture, downstream tasks, and evaluation metrics as seen in works like BRLTM \citep{meng2021brltm} and Hi-BEHRT \citep{li2022hibehrt} or use ICU-specific data \citep{zhang2022adadiag}. This heterogeneity complicates direct comparisons and makes it untractable to pinpoint the sources of improvement.
\begin{figure}[t]
	\centering
	\includegraphics[width=0.99\textwidth]{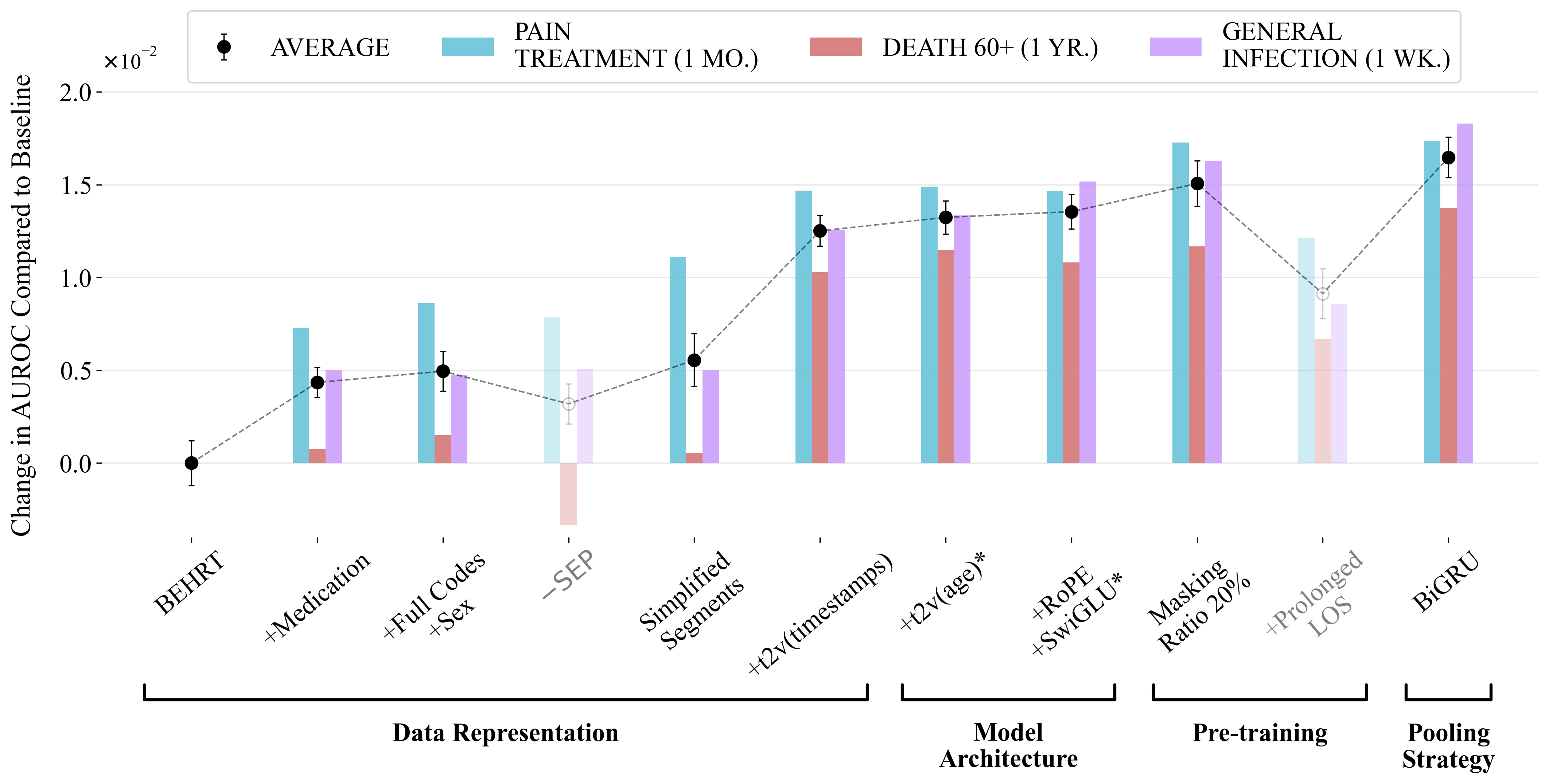}
	\caption{This figure depicts the incremental optimization of the configurations. Non-selected configurations are depicted in a lighter shade, indicating they were considered but not chosen based on their impact on the average Area Under the Receiver Operating Characteristic (AUROC), across three prediction tasks, whose prediction windows are denoted in parentheses. Models were trained in five-fold cross-validation and evaluated on one test set per task. An extension of these experiments for masking ratios and pooling strategies is demonstrated in \autoref{appendix:mr_and_heads_results}.\newline
 \small{* Model performance metrics are averaged not only across cross-validation folds but also over multiple pre-training runs to mitigate instabilities in convergence.}}
	\label{fig:optimization}
\end{figure}
\vspace{.5em}

\noindent The field of NLP quickly recognized that when optimally configured, base transformer models could outperform later variants \citep{yang2019xlnet}. A prime example is RoBERTa, which showed that an optimally configured base BERT model could achieve superior results \citep{liu2019roberta}. We suspect a similar situation exists in the EHR domain, where initial approaches like BEHRT and Med-BERT may not be fully optimized. \\ \indent
Despite the parallels with NLP, it is also important to acknowledge the distinct characteristics of EHR data, notably its strong temporal aspect, irregular time intervals, and the distinct distribution of sequence lengths \citep{rasmy2021medbert}. These disparities necessitate a comprehensive evaluation of the components that have shown success in NLP in the context of EHR. Furthermore, design choices take a weighted importance in healthcare compared to NLP, as even minor performance improvements can significantly impact patient outcomes. \vspace{.5em}

\noindent In contrast to fields like radiology, where deep learning models are increasingly finding their way into clinical practice \citep{yasaka2018deep_radiology, montagnon2020deep_radiology, lauritzen2023assessing}, there is a lack of trust towards EHR-based models, a cornerstone for clinical adoption \citep{asan2020trust}. Therefore, by providing insights into these models, we increase their trustworthiness and make a step towards the integration of BERT-based EHR models into clinical workflows.\\ \indent
Specifically, our study examines the fundamental design choices in data handling, modeling, and training processes for applying BERT-based models to EHR data, drawing inspiration from NLP optimizations. By examining model performance on a broad and diverse set of tasks, we provide robust and extensive evidence to solidify our recommendations. Furthermore, we investigate the variability in model performance to ensure that the models deployed in healthcare are optimized responsibly to inform life-critical decisions. 

\subsection*{Generalizable Insights about Machine Learning in the Context of Healthcare}
Weak and outdated foundations, coupled with the heterogeneity in newer models, pose challenges in making direct comparisons of BERT-based EHR models. This creates a lack of trust, a major hurdle to clinical adoption, and restricts these models' use to research settings.
We address these issues by conducting a careful optimization rigorously evaluated across a diverse set of tasks to increase insights into and confidence towards BERT-based EHR models. Our main contributions consist of:
\begin{itemize}
    \item Key insights in the data representation, model architecture, and training protocol of EHR modeling through careful optimization
    \item A comprehensive and rigorous evaluation across 25 diverse clinical prediction tasks showcasing the consistency of the optimizations
    \item A strongly optimized and comprehensively evaluated foundation serving as a stepping stone for wider adoption of BERT-based EHR models
\end{itemize}

%% file: sections/related_works.tex
\section{Related Work}
This section examines BERT's transformative impact on EHR analysis, illustrating its application through notable examples. We also extend our discussion to the developments related to BERT in the NLP field.

\subsection{BERT in EHR}
BEHRT \citep{li2020behrt}, serving as a proof-of-concept, was pre-trained on EHR data from 1.6 million patients, using sequences of patient events, along with age and visit details, and employed Caliber codes mapping ICD-10 (International Classification of Diseases 10th Revision) codes onto phenotypes. The model was evaluated in disease prediction tasks, showcasing superior performance over traditional recurrent neural network approaches. Their model used BERT's \citep{devlin2018bert} self-supervised masked language modeling (MLM) pre-training (PT) technique.\\ \indent
Another early example was G-BERT \citep{shang2019gbert}, which introduced a graph-based approach that leveraged GNNs to represent the hierarchical structure of the ontologies of diagnosis and medication codes. G-BERT was trained on MIMIC-III \citep{johnson2016mimic3}, a small but public dataset consisting of 40k patients from a critical care unit in the US. They evaluated using medication recommendation, a multilabel classification task recommending medications based on diagnosis codes.\\ \indent
To fully leverage the advantages of transformers in large-scale PT, Med-BERT \citep{rasmy2021medbert} scaled up the PT dataset to encompass over 28 million patients. It introduced a novel PT task, aimed at predicting extended hospital stays, and expanded its vocabulary to cover all ICD-9 and ICD-10 codes. Med-BERT's design also diverged from its predecessors by omitting age information and separator tokens and experimenting with various classification heads, among which the Bi-GRU \citep{schuster1997bidirectional} turned out to be the most effective.\\ \indent
Since then, BEHRT and Med-BERT have been the foundation of many subsequent models. 
For example, Hi-BEHRT \citep{li2022hibehrt} incorporated hierarchical attention mechanisms to better handle patients with extensive medical histories. 
CEHR-BERT \citep{pang2021cehr} integrated time intervals between patient visits to address the critical aspect of temporal information in EHRs, offering a more nuanced view of patient histories. 
Targeted-BEHRT \citep{rao2022targetedbehrt} performed causal inference on EHR data reproducing associations found in randomized controlled trials. 
BRLTM \citep{meng2021brltm} included medications, procedures, gender, and topics from clinical notes for depression and chronic disease prediction.
RareBERT \citep{prakash2021rarebert} extended Med-BERT with an adaptive loss for class imbalance, type embeddings, and temporal reference embeddings to better handle rare diseases. 
\subsection{Transformers in NLP}
Despite the advancements on EHR data, the journey of transformer-based models in healthcare mirrors the early challenges faced by foundational models like BERT \citep{devlin2018bert}. Initial iterations were soon recognized to be underoptimized, brought to light by replication studies such as RoBERTa \citep{liu2019roberta}.\\ \indent
Their study showed that the original BERT model was under-trained and questioned the usefulness of its Next Sentence Prediction task. They advocated for dynamic over static masking, including longer sequences, and larger batch sizes. They highlighted the importance of key design choices by showcasing better performance than later variants like XLNet \citep{yang2019xlnet}.\\ \indent
We argue that this critique extends to the realm of BERT-based EHR models, which exhibit a wide range of configurations. The diversity in vocabularies, datasets, tasks, and incorporated clinical information underscores the challenge of isolating the effect of individual model components. This heterogeneity complicates the assessment of progress within the field, blurring the lines between genuine innovation and incremental modifications.\\ \noindent
Lastly, transformer-based models have seen rapid development, primarily driven by the rise of Large Language Models (LLMs). Innovations include new activation functions, \citep{shazeer2020glu}, improved position embeddings \citep{su2023roformer}, and compute efficient fine-tuning methods \citep{dettmers2023qlora}. Furthermore, comprehensive advancements such as the transformer++ recipe have been integrated into modern LLMs like PaLM \citep{chowdhery2022palm}, and Llama \citep{touvron2023llama}, marking a significant shift from earlier architectures like BERT. 

%% file: sections/methods.tex
\section{Methods}
\begin{figure}[t]
	\centering
	\makebox[\textwidth][c]{\includegraphics[width=\linewidth]{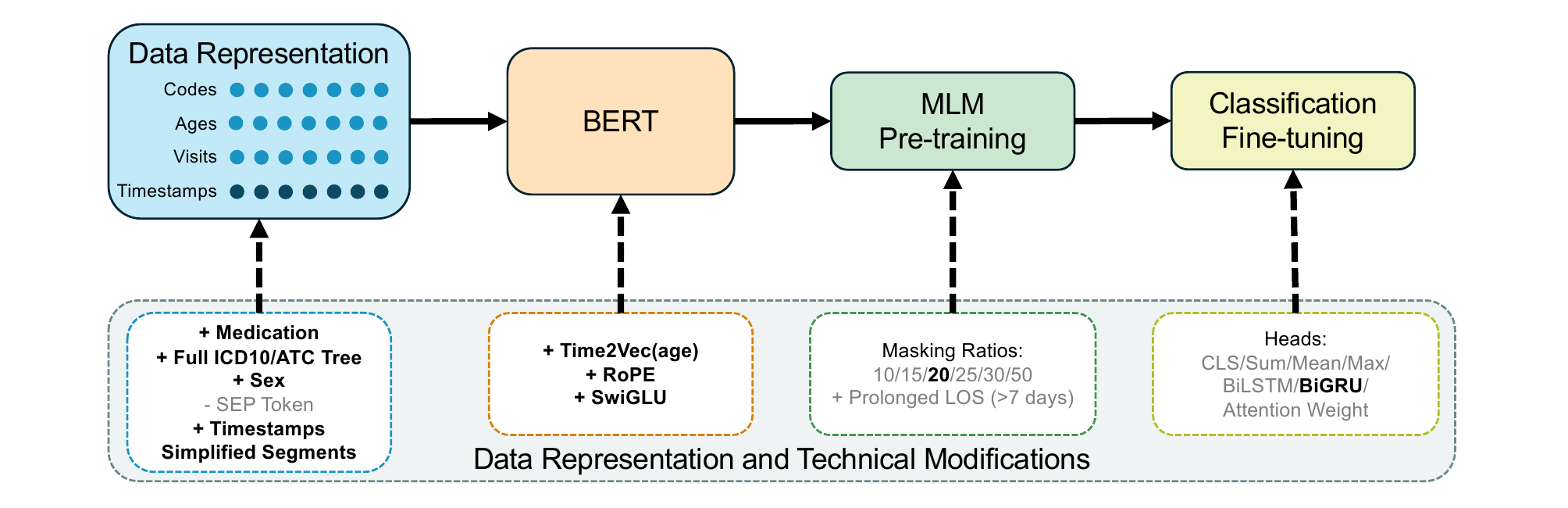}}
        \vspace{-1em}
	\caption{Overview of the iterative optimization procedure, highlighting adapted settings in bold. Symbols denote either addition (+) or removal (-) of a setting with no symbol indicating a direct change. We augmented the input data with medication codes, patient sex, event timestamps, full International Classification of Diseases (ICD-10), and Anatomical Therapeutic Chemical (ATC) codes. Improvements to the model include integrating time2vec embeddings, Rotary Position Embeddings (RoPE), and the SwiGLU activation function. For pre-training (PT), we investigated various masking ratios and Med-BERT's secondary PT task: predicting prolonged length of stay in the hospital. We investigated various pooling strategies for fine-tuning (FT), finalizing on Bidirectional Gated Recurrent Units (BiGRU) based on its superior results.}
	\label{fig:main_methods_figure}
\end{figure}
We will introduce our base model and the modifications tested, separated into a data presentation and a technical part. We test a few representative components in different areas, covering as much ground as possible while keeping the computational demand within our budget. 
The data representation part investigates the impact of representing commonly available background EHR data with little to no implementation overhead. The technical part investigates the impact of improved architecture, embeddings, and training schemes, primarily relevant to machine learning practitioners. Modifications are introduced incrementally, evaluating their impact on the performance of three generic downstream tasks and including them when there is an increase in average Area Under the Receiver Operating Characteristic (AUROC). An overview of this procedure is shown in \autoref{fig:main_methods_figure}. Next, we describe the optimization procedure before defining the various downstream tasks and close with the hyperparameters of our architecture and optimization. 

\subsection{Base Model}
We used a slightly adapted BEHRT as our starting point. BEHRT's input data consist of medical events with CLS (classification) and SEP (separator) tokens, age in years, and visits denoted by "positions" and binary visit encodings (segments). \\ \indent
In BEHRT, the events consist of level four ICD10 codes alongside Read codes \citep{booth1994read}, mapped to 301 Caliber codes \citep{kuan2019caliber} representing disease phenotypes. Lacking access to Read codes, we utilized level four ICD10 codes. Similarly, we employed level four ATC (Anatomical Therapeutic Chemical) codes for experiments involving medications. We believe this also strengthens the generalization of our experiments, as ICD and ATC are more commonly used than mappings such as Read and Caliber codes.\\ \indent
We adopted Med-BERT's leaner configuration for hyperparameters to maintain manageable model dimensions, as BEHRT could use more layers and heads due to its smaller vocabulary size. Further details are provided in \autoref{sec:hyperparameters}. \\ \indent
Henceforth, we will refer to our modified baseline configuration simply as BEHRT for ease of discussion.

\subsection{Data Representation}
We investigate different key areas of data representation by including more data sources, extending existing data, and exploring visit representation. This approach provides valuable insight into the aspects of data representation that are crucial for performance gains. 
\paragraph{Medication Codes}
We adopted the BEHRT model to include medication codes, recognizing their potential to improve predictive modeling \citep{wei2016combining_ehr}. Although the inclusion of medication codes, due to their similar coding system to diagnoses, is not unconventional, it has yet to become standard practice. We consider medication a vital addition to the model, as diagnosis codes are often incomplete or inadequate \citep{horsky2017accuracy_icd10}. 
For simplicity and computational feasibility, our current model includes only the medication codes, akin to how we handle diagnoses, without considering dosage information. Although our dataset also contains lab tests, vital signs, and procedures, with significantly higher event counts (672.2 million, 605.9 million, and 300.3 million events, respectively), we focus solely on the 227.6 million medication events due to their direct relevance to treatment decisions.

\paragraph{Full-Depth Codes and Background Sentence}
We propose including more fine-grained data by using the full-depth medical and diagnostic codes and including the patient's sex. It is common for EHR models to limit the input to higher-level codes, such as in BEHRT \citep{li2020behrt}. This can be beneficial as high-level representation decreases the vocabulary size which in turn decreases model size. It may also reduce noise to represent information on a coarse level only. Despite these advantages, it can also remove critical information and might especially be detrimental when rare conditions are of interest. BEHRT mapped their codes to 301 Caliber categories, far less than the number of ICD10 codes. When utilizing level four codes, our model works with a vocabulary size of 2,475, whereas using the full codes expands the vocabulary to 17,469. This highlights the balance between model complexity and the granularity of information captured. \\ \indent
To include the patient's sex in the model, we prepend a token representing male or female to their sequence. We foresee that other relevant demographic variables could also be included to form a 'background sentence'. 

\paragraph{Separator Tokens}
Separator (SEP) tokens are designed to provide an additional signal to separate visits. Since visit numbers are given explicitly to the model, Med-BERT hypothesized that removing SEP tokens should not diminish performance. Furthermore, for sequences longer than the truncation length, removing SEP tokens allows for including more medical events in the sequence.

\paragraph{Simplified Segments}
We believe that BEHRT has redundant visit separation (segment) information, including separator tokens, visit numbers, and binary visit encodings. We hypothesize that the model can infer the binary visit encodings through visit numbers and differentiate between visits in this way. We remove the binary visit embedding and change BEHRT's initialization of the visit number embedding to a standard one rather than their cosine and sinus version.

\paragraph{Timestamps}
We suggest the inclusion of medical event timestamps to enhance our model, moving beyond BEHRT's sole reliance on age for a notion of time. Timestamps allow for the precise tracking of short-term changes and are crucial for accurately predicting outcomes that depend on fine time resolution. Unlike the patient's age, which offers a \textit{relative} sense of time, timestamps provide an \textit{absolute} timeline, enabling the model to account for societal shifts, including pandemics, changes in treatment protocols, and seasonal trends. This feature is particularly useful for detecting shifts in clinical EHR coding practices and adapting to variable baseline risks in conditions like influenza, which intensifies in winter. 
\\ \indent
Furthermore, external factors can affect hospital interactions, as highlighted during the COVID pandemic when many avoided hospital visits, even for severe conditions \citep{nourazari2021decreased}. This shift can alter the interpretation of EHR sequences. We can account for these temporal variations in EHR semantics by incorporating event timestamps, ensuring our model remains accurate over time. Timestamps are input into the model as the time difference in hours from a specific reference point (January 26, 2020).

\subsection{Technical Modifications} \label{sec:technical}
Here, we introduce the technical components selected from three key technical areas: model architecture, pre-training, and finetuning strategies. The components have been chosen due to their presence and success in the EHR or NLP literature.

\paragraph{Time2Vec}
Like \citep{pang2021cehr}, we suggest substituting BEHRT's conventional age embedding with a Time2Vec \citep{kazemi2019time2vec} embedding. BEHRT's current model employs embeddings that rely on discrete age in years, which can be problematic, especially in pediatric cases where changes occur on shorter timescales. These embeddings may also struggle to discern the relationship between different ages. Time2Vec models this relationship explicitly, using differently parameterized periodic functions, by giving similar representations to similar ages. 
Adequately understanding the relationships between ages necessitates exposure to a wide range of examples for each age. To overcome these limitations and to accommodate the timestamps mentioned previously, we propose the adoption of Time2Vec. We leave the details of Time2vec and our implementation of it in \autoref{appendix:time2vec}.

\paragraph{Improved Transformer Recipe}
Numerous refinements have been suggested since the original transformer paper, and we incorporate two such advancements from the improved transformer recipe, dubbed transformer++, that are particularly effective in boosting performance: Rotary Position Embedding (RoPE) \citep{su2023roformer} and the SwiGLU activation function \citep{shazeer2020glu}. \\ \indent
RoPE replaces the standard positional embedding by encoding the absolute position through a rotation matrix and incorporating relative position in self-attention. The idea is to rotate the embeddings with an angle corresponding to the positional index of the event which is applied at every key and query multiplication in each attention head.\\ \indent
The SwiGLU activation function is a combination of Swish \citep{ramachandran2017swish}, also called Sigmoid Linear Unit (SiLU), and GLU \citep{dauphin2017glu}. It was shown to improve performance \citep{shazeer2020glu} and is defined as:
\begin{align}
    \textrm{SwiGLU}(x) &= (\textrm{Swish}(xW_1) \otimes xW_2)W_3\,, \\
    \textrm{Swish}(x) &= x * \mathrm{sigmoid}(x)
\end{align}

\paragraph{Pre-training Objective}
The training scheme of BEHRT plays a major role in the performance, as the strength of these models comes largely from PT, specifically Masked Language Modelling (MLM). Here, BEHRT copied the setup from BERT, masking 15\% of the tokens — referred to as the masking ratio (MR) —using the 80-10-10 strategy (80\% mask, 10\% random word, 10\% unchanged). We have yet to see any evidence-based reason for picking 15\% in EHR, and other modalities (images, audio, and videos) have much higher MRs \citep{dosovitskiy2021vit, he2021mae, baevski2020wav2vec2.0, tong2022videomae}. It has also been seen that larger models tend to favor larger MRs \citep{wettig2023mask}.

Furthermore, we tested the second binary classification PT objective introduced by Med-BERT \citep{rasmy2021medbert}, in which a prediction is made on whether the patient has a prolonged length of stay in the hospital ($>7$ days) for each patient. This task was run alongside MLM.

\paragraph{Pooling Strategies}
Lastly, we test different pooling strategies for the downstream tasks. It is common to use the CLS token, i.e., the first token in the sequence, as BERT introduced it as a special token specifically for classification. We investigate three additional simple pooling strategies, sum, mean, and max, which aggregate information across the sequence dimension. We also test a weighted variation of mean pooling, which uses an attention layer with a key and query to find the weights of tokens. Lastly, we test two bidirectional Recurrent Neural Networks (RNNs), BiGRU and BiLSTM, as this has been shown to increase performance \citep{rasmy2021medbert}.

\subsection{Optimization Procedure}\label{sec:optimization_procedure}
After introducing the various modifications, we now outline the optimization procedure in this section, as detailed in \autoref{fig:main_methods_figure}. Our approach begins with improving the data representation, followed by model improvements, and concludes with optimizing the PT process and fine-tuning (FT) pooling strategy. \\ \noindent
Each setting is introduced incrementally in the optimization procedure. We begin with the baseline model (BEHRT) and first introduce medication codes. If this modification enhances performance, it is incorporated into the model. Next, we test the second modification (Full ICD10/ATC Tree) on the updated model, which now includes the baseline and the improvement. This iterative process continues for each subsequent modification, with each new change tested on a model that incorporates all previous beneficial changes. \\ \noindent
For evaluation, we use three generic downstream tasks for generalizable performance: one-year death prediction for patients older than 60 y.o., one-month pain treatment prediction, and one-week infection prediction; the corresponding definitions are given in \autoref{appendix:disease_definitions}, consisting of diagnosis and medication codes as well as free text. 
Death encapsulates the final outcome and can signal the overall health status, encompassing various conditions and factors affecting patient well-being.
Pain treatment is very frequent and covers a wide range of causes. It can be both mild and severe, indicative of many underlying conditions, and thus quite general. 
Infection points to a compromised immune system, which also tells us about the general condition of a patient. \\ \indent
Our three tasks constitute a generalizable evaluation by covering a broad range of health conditions.
To compare the performance between models, we mainly considered the area under the receiver operating curve (AUROC) but also reported the area under the precision-recall curve (AUPRC) in \autoref{appendix:initial_experiment}. For each downstream task, we perform five-fold cross-validation (CV) evaluating on a test set and compute the mean and standard deviation.
The downstream task was a binary classification and was constructed as follows. For patients with positive outcomes, we used the outcome of interest as the index date. We then excluded all events within a specified prediction window leading up to the index date, with the window tailored to each downstream task. Disease definitions and prediction windows are given by expert clinicians, ensuring that the model sees no outcome-related information during prediction, preventing any information leakage. For patients with negative outcomes, we assign random censoring dates matching the distribution of the positive cases to avoid information leakage introduced by censoring.  This makes the censoring identical for both cases and controls. 
To account for the imbalance in the dataset, we sampled with the inverse square root of the frequency of each class \citep{mahajan2018sqrt, mikolov2013nlpsqrt}.

\subsection{Assessing Model Generalizability}
This section outlines our methodology for evaluating the model's performance across a diverse array of health conditions.
Our goal was to ensure the generalizability of our conclusions by covering the prediction of a wide spectrum of diseases and conditions (used interchangeably from now on), showcasing the optimized model's adaptability and precision compared to the baseline model in various clinical contexts.\\ \indent
Our comprehensive evaluation encompassed conditions with varying prevalence, from commonly prescribed treatments like pain medications to less frequent conditions such as pancreatic cancer. We considered how conditions develop over time ranging from chronic diseases like diabetes to acute events like major bleeding.\\ \indent
We also covered conditions with different associations to overall health and underlying factors. For instance, the strong correlation between stroke and overall health \citep{arboix2015cardiovascular} contrasts with the more complex interplay of genetic and environmental factors in schizophrenia \citep{stilo2019non_genetic_schizophrenia, henriksen2017genetics_schizophrenia}, which presents a subtler EHR data signature. We also covered conditions with a limited range of causes, such as anaphylaxis, which primarily involves drug-related causes, and those that can have a wide range of underlying conditions, such as sleep disorder. The conditions are listed in \autoref{tab:generalization_cohort_stats} together with prediction window and age groups, which were chosen based on relevance and considering the age distributions provided in \autoref{appendix:age_distribution}.\\ \indent
This structured approach to task selection is designed to demonstrate the generalizability of our model in identifying a wide range of health outcomes, underscoring its potential utility in diverse clinical scenarios without having to revisit the model choices for every new downstream task. We construct each downstream task as outlined in the previous \autoref{sec:optimization_procedure}, reusing the definitions from \autoref{appendix:disease_definitions}, and perform hypothesis testing as described in \autoref{appendix:p_value}.\\ \indent
In addition, we performed experiments using out-of-time evaluation to mimic real-world scenarios more closely. Here, we look at temporally separated outcomes for train, validation, and test sets. A detailed description is provided in \autoref{appendix:out_of_time}. This method allows the model to predict the future and assess its response to temporal shifts.

\subsection{Hyperparameters}\label{sec:hyperparameters}
To ensure a fair comparison, we standardized the hyperparameters across all experiments. We aimed to facilitate stable model convergence across all models while keeping manageable training durations. We utilized a base model architecture consisting of six layers and six attention heads with a hidden size of 192 and an intermediate size of 64. Sequences were truncated to 512 tokens, and a batch size of 256 for PT and 512 for FT was used. Learning rates were set to $10^{-3}$ for PT and $2\cdot10^{-5}$ for FT using a linear warmup of five epochs. Models were trained for 50 epochs during optimization and 150 epochs for the generalization experiments with an early stopping of five epochs. A more detailed overview can be found in \autoref{appendix:hyperparameters}.

%% file: sections/cohort.tex
\section{Cohort}
The data consists of EHRs from 1,825 million patients, prospectively recorded during the period from 2016 to the end of 2022 in the Capital Region of Denmark. We consider only patients with at least one code eligible, which results in 1,805 million patients. For FT, we deem at least one code to be a proper input for binary classification but use patients with $\geq 3$ codes for PT to ensure a proper input for the MLM objective task. An overview of the selection pipeline can be seen in \autoref{appendix:cohort_selection}.
\input{tables/initial_experiments_cohort}

\subsection{Cohort Selection}
For the optimization experiments, we randomly selected 500 thousand patients, split 80/20 into train and validation sets, and reserved 260 thousand for CV. The specific statistics for each CV task can be seen in \autoref{tab:optimization_cohort_stats}. This subgroup was used as part of the PT group for the final generalizability evaluation, where we used a total of 1,202 million patients, with the remaining 702 thousand used for CV. \\ \indent
We observed a discrepancy in prevalence rates between the optimization and generalization experiments. This could have been caused by the assignment of the subgroup to the generalization PT group, as the subgroup was filtered, which may have left fewer positive cases for the generalization evaluation. \\ \indent
To test generalizability across a wide range of tasks while managing computational demands, we maintained a manageable cohort size for the main experiments. However, we also performed "maximizing performance" experiments, where we extended the cohorts, including PT patients in the FT set for select tasks. Details of these tasks and the expanded cohort sizes are provided in \autoref{appendix:max_performance}.\\ \indent
The out-of-time evaluation, aimed at mimicking a real-world setting, was constructed by censoring training data on 01/01/2020. Without performing patient splits and keeping patients with at least one event we obtain 1,033,465 PT patients. The time separation is described in detail in \autoref{appendix:out_of_time} and patient numbers are provided in \autoref{tab:out_of_time}.\\ \indent
For model selection, we require each patient in the PT cohort to have at least three \underline{diagnoses} to ensure a fair comparison between the diagnosis only and the diagnosis+medication models. For the final evaluation, we require at least three of \underline{any} (diagnosis or medication) codes, to accurately reflect the increased cohort when including medication. \\ \indent
For each condition, we defined a relevant age and/or sex and a prediction window that we use for censoring, given in \autoref{tab:optimization_cohort_stats} for the optimization experiments and in \autoref{tab:generalization_cohort_stats} for the generalization assessment. The full age distributions for the cohorts before age group selection are given in \autoref{appendix:age_distribution}.
We consider patients to be positive when having at least one of the codes given in \autoref{appendix:disease_definitions} and negative otherwise. We address imbalance and potential information leakage using the methods described in \autoref{sec:optimization_procedure}.

\subsection{Data Extraction}
We extract all medical events, with an event consisting of three main variables: concept, timestamp, and admission ID. These events undergo minimal data processing through a simple three-stage process of replacing missing ICD10 codes with textual descriptions, attempting to impute missing values, only when confident, before excluding any events with missing info or ages outside a normal range. We leave a more detailed description in \autoref{appendix:dataprocessing}.

\subsection{Feature Choices} 
After processing, we end up with a simple data format that describes each medical event with the three main variables: concept, timestamp, and admission ID. 
The concept of each event is the medical code (diagnosis or medication) and is the main input to the model. 
We include the timestamp of the event and use it to calculate the patient's age. 
The admission ID, together with the timestamp, is used to calculate the visit numbers.

%% file: tables/initial_experiments_cohort.tex
\begin{table}[t]
\caption{Statistics for the different train-val sets for death (60+), pain treatment, and general infection for the optimization experiment. We provide the prediction window and mean age at censoring as well as the prevalence of positive patients.}
\centering
\begin{tabular}{
p{0.2\linewidth}
>{\centering\arraybackslash}p{0.144\linewidth}
>{\centering\arraybackslash}p{0.075\linewidth}
>{\raggedleft\arraybackslash}p{0.09\linewidth}
>{\raggedleft\arraybackslash}p{0.09\linewidth}
}
\toprule
\multicolumn{1}{p{0.2\linewidth}}{Condition \newline (Age Group)} & 
\multicolumn{1}{>{\centering\arraybackslash}p{0.144\linewidth}}{Prediction\newline Window} & 
\multicolumn{1}{>{\centering\arraybackslash}p{0.075\linewidth}}{Mean\newline Age} & 
\multicolumn{1}{>{\centering\arraybackslash}p{0.125\linewidth}}{Patients\newline[thousands]} & 
\multicolumn{1}{>{\centering\arraybackslash}p{0.125\linewidth}}{Prevalence\newline[\%]}
\\ \midrule
Death (60+) & 1 yr. & 75 & 55.5 & 25.6 \\
Pain Treatment & 1 mo. & 44 & 107.3 & 52.7 \\
General Infection & 7 d. & 49 & 106.4 & 57.7 \\
\bottomrule
\end{tabular}\label{tab:optimization_cohort_stats}
\end{table}

%% file: sections/results.tex
\section{Results on Real Data} 
\subsection{Evaluation Approach/Study Design} 
We first evaluated different components of data representation and technical modifications by pre-training for 50 epochs using a cohort of 500 thousand patients before doing a five-fold CV on death, pain treatment, and infection prediction. We iteratively added components to the pipeline if they improved the average AUROC across these three downstream tasks. \\ \indent
Subsequently, we compared the baseline (BEHRT) against a model with only improved data representation (\behrtdata) and a model with both improved data representation and technical components (\behrtbest). We evaluated the performance on diverse tasks to ensure the generalizability of the performance gains.
\vspace{.5em}

\noindent Before setting up our experiments, we investigated the variance in PT and FT by running the base model multiple times. We found that both yielded stable performances with standard deviations of $6.6\cdot10^{-4}$ and $3.2\cdot10^{-4}$ for PT validation loss and FT validation AUROC, respectively. We investigated the source of variance in FT by conducting a leave-2-out CV; see \autoref{appendix:leave_2_out} for details. The analysis revealed that the mean difference within test sets is $7.9\cdot10^{-4}$ and pairwise difference for the means across test sets is $4.1\cdot10^{-3}\,$, showcasing that variation is mainly attributed to the test set.
These results gave us sufficient confidence to justify a single PT, followed by a five-fold CV. To ensure robust evaluation, we intentionally used larger-than-usual test sets. 

\subsection{Optimization Experiments}
Through iterative enhancements in data representation, model architecture, and training as visualized in \autoref{fig:optimization}, our optimization experiments have systematically advanced AUROC across multiple clinical prediction tasks.
\paragraph{Data Representation} The proposed data changes outline two main contributions, medication, and timestamps. Both yielded a clear and significant increase in AUROC, whereas full-length codes and sex showed a smaller increase. This makes it clear that adding deeper and more precise data has a strong positive impact on the model. \\ \indent
However, it must be mentioned that adding medication introduced 258 new codes, which increased average sequence lengths by 143.3 codes and training time by five times compared to baseline. Changing to full-depth codes and adding sex introduced almost 15 thousand new codes, which increased the model size by 3.7 times up to 7.933 million parameters. This is due to the increased size of the embedding layer, which increases training time, as these are trainable parameters but do not affect inference time. \\ \indent
Notably, removing SEP tokens worsened performance, primarily for death, while the simplified segments yielded improvements. This reveals that the model mainly uses the SEP tokens to differentiate between visits and relies on explicit embeddings. 

\paragraph{Technical Components} The two technical components for the model, Time2Vec for age and the improved transformer recipe, yield smaller AUROC gains.
We noticed during PT that RoPE, which is part of the improved transformer recipe, yielded a huge boost in MLM performance ($33\%\rightarrow58\%$ for top-1 and $82\%\rightarrow91\%$ for top-10 accuracy). This large improvement primarily comes from an improved ability to predict masked medication codes $(35\%\rightarrow61\%)$ with a smaller improvement for diagnosis codes $(20\%\rightarrow22\%)$.

\paragraph{Masking Ratios} We found that an MR of 20\% was optimal, showing higher AUROCs for each of the three tasks compared to the base MR of 15\%, as seen in \autoref{appendix:mr_and_heads_results}. At the same time, we observed that changes in masking ratios did not result in substantially different average AUROCs, and in general, the impact was inconsistent across the downstream tasks.
\paragraph{Pooling Strategies} The largest improvement in pooling strategies comes from the BiGRU head, but we show in \autoref{appendix:mr_and_heads_results} that most techniques performed better than the standard CLS token. Even simple aggregation functions, such as mean and max pooling, which use no additional trainable parameters, showed general improvements compared to CLS.
\input{tables/main_table_vMikkel}
\subsection{Generalization Results} 
Our analysis revealed that the \behrtbest\;model outperformed the baseline in all evaluated scenarios, as detailed in \autoref{tab:results}. When compared to \behrtdata, significant improvements were observed in only seven downstream tasks, notably including death, pain treatment, and general infection prediction. Despite this, the \behrtbest\: model consistently demonstrated increased average AUROC scores across the board with exceptions in only five tasks, with two of them being equal in performance. \\ \indent
It is worth noting that tasks where significance was not achieved for both \behrtdata and \behrtbest\:, compared to baseline, were predominantly among the more challenging ones, characterized by lower AUROC scores. Specifically, six out of eight tasks lacking significant improvements had AUROC scores below 0.65.
\vspace{0.5em}

\noindent The impact of increasing the CV set can be seen in \autoref{appendix:max_performance}. Breast cancer and preterm birth has notable AUROC increases, while stroke and schizophrenia remain largely unchanged despite the significant increases in patient numbers. This suggests a performance saturation at lower patient counts, supported by their high baseline AUROCs.

\subsection{Out-of-Time Evaluation}\label{sec:out_of_itme}
When mimicking a real-world setting by performing an out-of-time evaluation, the size of the train and test sets increases dramatically and the prevalence decreases. Most tasks see a substantial performance increase, while death sees a small decrease and stroke a larger decrease as detailed in \autoref{tab:out_of_time}. Overall, this showcases the model's ability to handle temporal shifts well.
\input{tables/real_world_results}

%% file: tables/main_table_vMikkel.tex
\begin{table}[ht!]
\centering
 \caption{Model performance for BEHRT (baseline), \behrtdata (BEHRT with optimized data representation) and \behrtbest\; for multiple conditions alongside average age at censoring, patient numbers, and prevalence for positive patients in the test set. If applicable, the selected age group and sex are given in parentheses.}
\begin{tabular}{
p{0.31\linewidth}
>{\centering\arraybackslash}p{0.144\linewidth}
>{\centering\arraybackslash}p{0.06\linewidth}
p{0.0\linewidth}
p{0.0\linewidth}
p{0.0\linewidth}
p{0.0\linewidth}
p{0.0\linewidth}
}
\toprule
Condition\newline(Age Group/Sex) & 
Prediction Window & 
Mean Age & 
\multicolumn{2}{>{\centering\arraybackslash}p{0.14\linewidth}}{Patients\newline [thousands]/\newline Prevalence[\%]} & 
\multicolumn{3}{>{\centering\arraybackslash}p{0.27\linewidth}}{AUROC [\%]\newline BEHRT/\behrtdata /\newline\behrtbest} \\
\midrule
\textbf{Cancer} \\
Basal Cell Carcinoma ($70+$) & 1 yr. & 80 & \dualcol{20.6}{4.7} & \multicol{64.4}{65.8**}{\textbf{67.2}**} \\
Breast ($60+$/f.) & 3 mo. & 74 & \dualcol{13.5}{4.4} & \multicol{57.0}{57.6}{\textbf{62.4}} \\
Colon ($60+$) & 3 mo. & 75 & \dualcol{35.5}{1.3} & \multicol{61.4}{\textbf{63.6}**}{62.8} \\
Lung ($60+$) & 3 mo. & 74 & \dualcol{35.0}{2.3} & \multicol{66.5}{67.9*}{\textbf{68.4}**} \\
Pancreatic ($60+$) & 3 mo. & 74 & \dualcol{37.7}{0.5} & \multicol{52.9}{\textbf{57.5}}{56.5} \\
\textbf{Cardiovascular}\\
Arrhythmia ($60+$) & 6 mo. & 76 & \dualcol{35.5}{10.7} & \multicol{68.9}{71.5***}{\textbf{71.6}***} \\
Myocardial Infarction ($60+$) & 14 d. & 75 & \dualcol{40.4}{2.6} & \multicol{65.7}{68.3***}{\textbf{68.7}***} \\
Stroke ($60+$) & 14 d. & 75 & \dualcol{30.2}{1.3} & \multicol{64.5}{69.4**}{\textbf{70.2}***} \\
\textbf{Chronic} \\
Alzheimer’s Disease ($70+$) & 3 mo. & 82 & \dualcol{23.1}{3.7} & \multicol{73.9}{76.5**}{\textbf{77.0}***} \\
Diabetes & 6 mo. & 65 & \dualcol{107.4}{4.9} & \multicol{83.5}{\textbf{84.8}***}{\textbf{84.8}***} \\
Osteonecrosis ($70+$) & 6 mo. & 77 & \dualcol{25.7}{0.2} & \multicol{82.5}{\textbf{84.6}}{84.2} \\
\textbf{Death} ($60+$) & 1 yr. & 80 & \dualcol{35.0}{22.7} & \multicol{86.6}{87.5***}{\textbf{87.7}***\dgr} \\
\textbf{Emergencies} \\
Anaphylaxis & 24 hr. & 32 & \dualcol{169.1}{0.2} & \multicol{71.0}{71.4}{\textbf{71.9}} \\
Diarrhea & 3 d. & 59 & \dualcol{158.1}{1.4} & \multicol{85.6}{86.5**}{\textbf{87.5}**\dgr} \\
Major Bleeding & 24 hr. & 59 & \dualcol{162.6}{4.6} & \multicol{78.2}{78.9**}{\textbf{79.7}***\dgr} \\
\textbf{Infections} \\
General Infections & 7 d. & 47 & \dualcol{141.0}{42.8} & \multicol{78.2}{79.7***}{\textbf{79.9}***\dgr} \\
Pneumonia & 14 d. & 61 & \dualcol{146.6}{5.5} & \multicol{82.3}{83.0***}{\textbf{83.2}***\dgr} \\
\textbf{Psychiatric} \\
Depression ($15-30$) & 3 mo. & 45 & \dualcol{142.7}{1.8} & \multicol{71.3}{71.8**}{\textbf{71.9}**} \\
Schizophrenia ($20-30$) & 3 mo. & 24 & \dualcol{12.7}{1.2} & \multicol{83.8}{85.8**}{\textbf{86.0}**} \\
\textbf{Pain treatment} & 1 mo. & 47 & \dualcol{133.8}{42.0} & \multicol{80.8}{82.1***}{\textbf{82.2}***\dgr\dgr} \\
\textbf{Pregnancy-Related} \\
Preterm birth & at pregnancy & 30 & \dualcol{11.1}{13.7} & \multicol{53.4}{54.5}{\textbf{59.5}} \\
Fetal Growth Restriction & at pregnancy & 31 & \dualcol{11.1}{6.9} & \multicol{56.9}{61.6***}{\textbf{61.7}*\dgr\dgr\dgr} \\
Placental insufficiency & at pregnancy & 32 & \dualcol{11.1}{0.5} & \multicol{\textbf{58.8}}{56.3}{58.6} \\
Large fetus & at pregnancy & 31 & \dualcol{11.1}{2.4} & \multicol{57.9}{60.8}{\textbf{61.6}} \\
\textbf{Others} \\
Sleep disorder ($20-40$) & 3 mo. & 30 & \dualcol{34.0}{4.1} & \multicol{74.8}{\textbf{75.9**}}{\textbf{75.9}**} \\

\bottomrule
\multicolumn{8}{@{}l}{\small{$^{*/**/***}\,$ Model is significantly better than BEHRT with $p<0.05/0.01/0.001$}}\\
\multicolumn{8}{@{}l}{\small{\dgr\textsuperscript{/}\dgr\dgr\textsuperscript{/}\dgr\dgr\dgr\, Model is significantly better than \behrtdata with $p<0.05/0.01/0.001$}}
\end{tabular}\label{tab:results}
\end{table}
\clearpage

%% file: tables/real_world_results.tex
\begin{table}
\caption{Results obtained mimicking a real-world setting, performing out-of-time evaluation, with train data being censored at 01/2020 and the test set contains outcomes from 09/2021 onwards. For comparison, we provide the difference in mean AUROC to the generalization results from \autoref{tab:results}.}
\label{tab:out_of_time}
\begin{tabular}{>{\raggedright\arraybackslash}p{0.26\linewidth} >{\centering\arraybackslash}p{0.18\linewidth} >{\centering\arraybackslash}p{0.12\linewidth} >{\centering\arraybackslash}p{0.15\linewidth}>{\centering\arraybackslash}p{0.15\linewidth}}
\toprule
Condition\newline(Age Group/Sex) & Patients train/test [thousands] & Prevalence in test [\%] & AUROC [\%] & $\Delta$ AUROC [\%]\\
\midrule
Breast Cancer ($60+$/f.) & 72.0/262.9 & 0.9 &64.4&$+2.0$ \\
Stroke ($60+$) & 174.3/516.8 & 0.2 & 67.1 & $-3.1$ \\
Diabetes & 593.9/1,657.1 & 0.9 & 85.6 & $+0.8$ \\
Death ($60+$) & 281.3/441.0 & 9.3 & 87.4 & $-0.3$\\
Schizophrenia ($20-30$) & 59.0/219.9 & 0.1 & 87.7 & $+1.7$\\

\bottomrule
\end{tabular}
\end{table}

%% file: sections/discussion.tex
\section{Discussion} \label{sec:discussion}
We have carefully optimized a BEHRT on three generic tasks, followed by a rigorous evaluation across a broad spectrum of clinical prediction tasks, showcasing the generality of our findings. This not only lays a solid foundation for future research but also sets a benchmark for the meticulous evaluation necessary to develop effective healthcare models.
\vspace{0.5em}

\noindent 
The discrepancy in performance gains between the MLM and our downstream EHR tasks, observed with the improved transformer recipe, could be explained by the fact that EHR tasks are not as closely tied to the PT objective as in NLP. EHR tasks that are more closely related to MLM might minimize the discrepancy. This might also explain our observation that different masking ratios impact different tasks in an individual and inconsistent manner. Developing a more tailored PT objective, that closely aligns with the nature of EHR downstream tasks, might alleviate this discrepancy altogether. 
\vspace{0.5em}

\noindent The generalization results show a clear trend. Improving data representation consistently, except for placental insufficiency, improves performance. However, expanding the data representation can be associated with increased model size and training times. Currently, EHR models are small compared to other domains, so increases are still manageable with a low compute budget. The technical components give the models an additional performance boost across all but four tasks and reach significance in seven of them. The technical components also never perform significantly worse and do not affect training times much, making them a solid addition to the optimized model. 
\vspace{0.5em}

\noindent
Increasing the FT set size yielded limited improvements for some tasks, suggesting performance saturation beyond a certain patient number. This saturation occurs faster with higher base scores, as seen with schizophrenia. Thus, adding more patients may not always help; instead, focusing on data representation and technical components could be more effective. \\ \noindent
The out-of-time evaluation shows the model's robustness to temporal shifts, maintaining high scores even with unseen time periods, likely due to the time2vec embedding. We believe this method is appropriate, as it mirrors real-world scenarios where models are used without constant retraining. The performance decrease on stroke might be caused by the larger prediction window with a minimum of three months, while the generalization experiment was performed with a 14-day prediction window.
\vspace{0.5em}

\noindent Here, we highlight the performance of our model on key examples to illustrate its comparative strengths. In stroke prediction, our model achieves state-of-the-art results with an AUROC of 70.7\%, which exceeds the 66.9\% AUROC reported in recent studies \citep{teoh2018towards_stroke_prediction}. It also shows comparable performance in preterm birth prediction with an AUROC of 64\% versus the 65\% reported in the literature \citep{alsaad2022predict_preterm_birth}, and in breast cancer prediction, where it achieves an AUROC of 63.9\% nearly matching the 64.8\% in existing studies \citep{wu2017breast_cancer}. 
Notably, we used narrower age windows in our work for stroke and breast cancer and employed fewer modalities for preterm birth. These examples illustrate our model’s strengths, even under more constrained conditions.
\vspace{0.5em}

\noindent Our optimization experiments were conducted on three distinct tasks, introducing some potential bias. The three out of seven tasks where \behrtbest\; significantly outperforms \behrtdata\; (BEHRT with optimized data representation) constituted the optimization tasks. This indicates a possibility for bias during the optimization, but we designed the tasks to be general and diverse to mitigate this effect.

\paragraph{Limitations}
Due to the sensitive nature of EHR data, it can be problematic to replicate setups exactly as public large-scale EHR datasets are rare. 
We thus used BEHRT with hyperparameters adapted to our dataset, which is also not publicly available. 
We also did not convert our codes into Caliber codes, which we believe hold greater value since ICD10 and ATC codes are more widely employed. 
\vspace{0.5em}

\noindent We found that the data representation and especially the inclusion of medication greatly impacted performance. Despite this, we only investigated the effect of medication codes and ignored sources like lab tests, vital signs, and procedure codes. These data sources are not as easy to integrate as medications and introduce many more events. 
This made it not only hard to integrate all data sources but also computationally infeasible for us to run a comprehensive optimization and evaluation. 
\vspace{0.5em}

\noindent
A key limitation of our work is the data size and collection period. Our cohort consists of approximately 1.8 million patients, significantly smaller than the dataset used for Med-BERT and tiny compared to typical NLP data sets. The data spans only a seven-year window, limiting the number of diagnosis and medication codes, as well as the feasible prediction windows. 
Additionally, EHR data quality varies as delays in code registration or missing codes can occur, potentially leading to incorrect patient labeling.
Although we partially address this by including free text descriptions, the issue of missing codes remains. 
Longer prediction windows would be preferable to avoid potential information leakage, especially for slowly developing conditions such as cancer. However, extending these windows proves challenging due to the short data collection period.
\vspace{0.5em}

\noindent Our study focuses only on BERT-based EHR models, ignoring the classical models like Linear Regression, Random Forest, and Gradient Boosted Trees (XGBoost). While BERT-based EHR models have proven to beat simpler models, they have yet to be tested vs. XGBoost. We simply focus on BERT-based models as we believe these models have the most potential due to their versatility and scalability, as evident in NLP and computer vision. These models also have the potential to be multimodal in the future, as seen with LLMs, making them a strong candidate model for the future.
\vspace{0.5em}

\noindent We show that improved data representation significantly boosts performance, with technical components providing an additional improvement. Although we only tested some modifications, we hope they act as representative examples and provide valuable insights into their respective areas. The consistent effects of these modifications remain unexplained, highlighting an important area for future research to uncover and analyze the underlying patterns.

%% file: sections/appendix.tex
\section{Time2vec Implementation}\label{appendix:time2vec}
Given a scalar input $\tau$ Time2Vec (t2v) produces a vector of size $k$ as follows:
\begin{equation}
    \mathbf{\mathrm{t2v}}(\tau)[i] = 
    \begin{cases}
        \omega_i \tau + \phi_i\,, & \text{ if i = 0}\,. \\
        \mathcal{F}(\omega_i \tau + \phi_i)\,, & \text{ if 1 $\leq$ i $\leq$ k}\,,
    \end{cases}
\end{equation}
where $\mathrm{t2v}(\tau)[i]$ is the $i^{th}$ element of $\mathrm{t2v}(\tau)$, $\mathcal{F}$ is the periodic function, where we choose cosine,
$\omega_i\,$ and $\phi_i$ are learnable frequency and phase-shift parameters. Time2vec uses fewer parameters than a standard embedding and can easily be implemented as a drop-in replacement. 

We noticed a certain degree of instability when using time2vec out of the box, but observed that scaling $\tau$ and clipping the $0^{th}$ dimension helped with this issue and stabilized the performance. After a rough hyperparameter search, we found stability when scaling age with a factor of $10^{-2}$, constraining it from 0 to around 1, and scaling absolute time with a factor of $10^{-4}$, converting a year of absolute time (8760 hours) to $\approx 1$. Furthermore, the $0^{th}$ dimension was clipped to $[-100, 100]$ to avoid large embedding values for very early events. 

\section{P-value Computation}\label{appendix:p_value}
To assess the statistical significance between the base and best models' performances, we calculated one-sided p-values using a t-test, under the null hypothesis that the model is not superior to either BEHRT or \behrtdata, if applicable. The means and standard deviations, based on five observations each, allowed for precise standard error estimation and subsequent t-statistic computation for each disease category. We apply the Benjamini-Hochberg procedure to account for the false discovery rate.

\section{Data processing}\label{appendix:dataprocessing}
\paragraph{Code-level replacement}
Some events had missing ICD10 codes, but also an accurate textual description of the diagnosis. Rather than excluding the events with missing ICD10 codes, we replace the missing value with the textual description. We account for these replacements when defining conditions, as each diagnosis code can now have multiple tokens to represent it. 

\paragraph{Admission-level Imputation}
Missing admission IDs and timestamps can sometimes be inferred by using surrounding events. Specifically, we try to impute admission IDs and timestamps, \textit{only} in the places where this can be done confidently. 

Missing admission IDs can be inferred when a missing admission ID exists between two identical admissions, e.g. [A None A]. Here, it is highly likely that the missing admission is part of the A admissions. Any missing admission ID that cannot be inferred is replaced with a unique identifier and treated as a standalone admission.

Some missing timestamps can be imputed if they are part of an admission with at least one other entry. Here, the missing timestamp can be estimated as most admissions lie within a similar time frame. The first, average, or last timestamp can be used, where we choose the last so as not to interfere with the ordering of the other codes. 

\paragraph{Event-level Exclusion}
Any event still containing missing values or events where the patient's age is outside a normal range ($0 \le \text{age} \le 120$) is removed.

\section{Hyperparameters}\label{appendix:hyperparameters}

\paragraph{Architecture Specification}
As part of our architecture choices, we adopted settings from Med-BERT with six layers, six attention heads, and a hidden dimension size of 192. These were chosen due to their efficient performance on larger datasets. The intermediate size is 64, with a context length of 512, considering variations in vocabulary sizes which affect model size.
\paragraph{Optimization and Training Parameters}
We used an epsilon value of $10^{-6}$ for the AdamW optimizer. The learning rates were set to $10^{-3}$ for PT and $2\cdot10^{-5}$ for FT, with a linear warmup phase of five epochs. Optimization experiments were limited to 50 epochs with an early stopping mechanism based on validation loss for PT and AUROC for FT. For the generalization experiments, the number of training epochs was increased to 150.
\paragraph{Computational Resources and Training Duration}
Training was conducted using Nvidia A100 GPUs with 80 GiB of VRAM. Batch sizes were set at 256 for PT and 512 for FT, ensuring maximal GPU utilization for our largest model versions. The duration for PT one model on 400,000 patients ranged from 2.5 to 16.5 hours, depending on model size.

\section{Leave-2-Out Cross Validation}\label{appendix:leave_2_out}
The main objective of the leave-2-out CV experiment was to analyze the sources of variance in model performance. The dataset was divided into five distinct folds, facilitating a comprehensive assessment through multiple test runs. During each iteration of the experiment, two of the five folds were selected: one served as the validation set and the other as the test set. This configuration allowed each test set to be evaluated by two different models, which were trained on overlapping subsets of the remaining three folds.
The experiment was structured to execute a total of 10 unique runs, each providing insights into model behavior under different training and validation scenarios. For each test set, we computed the performance differences between the two models and then averaged these differences to quantify the variability within the model evaluations. Furthermore, this average was compared against the mean of the pairwise differences in the mean performances across different test sets. This comparison aimed to distinguish between the intrinsic variability of the model and variability attributable to the composition of the test sets.
\clearpage

\section{Maximizing Performance}\label{appendix:max_performance}
\vspace{-.7em}
\begin{table}[H]
\caption{Results obtained with \behrtbest\:model for an increased cross-validation set on select conditions. AUROC is given with standard deviation computed from the five folds. The same test sets as described in \autoref{tab:results} were used.}
\label{tab:max_performance}
\begin{tabular}{>{\raggedright\arraybackslash}p{0.34\linewidth} >{\centering\arraybackslash}p{0.18\linewidth} >{\centering\arraybackslash}p{0.18\linewidth} >{\centering\arraybackslash}p{0.18\linewidth}}
\toprule
Condition\newline(Age Group/Sex) & Patients in CV [thousands] & Prevalence [\%] & AUROC [\%] \\
\midrule
Breast Cancer ($60+$/f.) & 124.3 & 5.2 & 63.9$\pm$0.3 \\
Preterm Birth & 100.0 & 13.7 & 63.7$\pm$0.2 \\
Schizophrenia ($20-30$) & 89.2 & 1.4 & 85.9$\pm$0.6 \\
Stroke ($60+$) & 276.2 & 1.5 & 70.7$\pm$0.5 \\
\bottomrule
\end{tabular}
\end{table}

The impact of increasing the CV set on model performance varies by condition, as shown in \autoref{tab:results}
and \autoref{tab:max_performance}. For breast cancer and preterm birth, AUROC increases by 2.4\% and 7.1\%, respectively. However, AUROCs for stroke and schizophrenia remain largely unchanged despite a 3.6-fold and 2.5-fold increase in patient numbers. This suggests performance saturation at lower patient counts, given their high baseline AUROCs of 70\% and 86\%.
\section{Out of Time Evaluation}\label{appendix:out_of_time}
\vspace{-.7em}
\begin{figure}[H]
	\centering
	\makebox[\textwidth][c]{\includegraphics[scale=0.5]{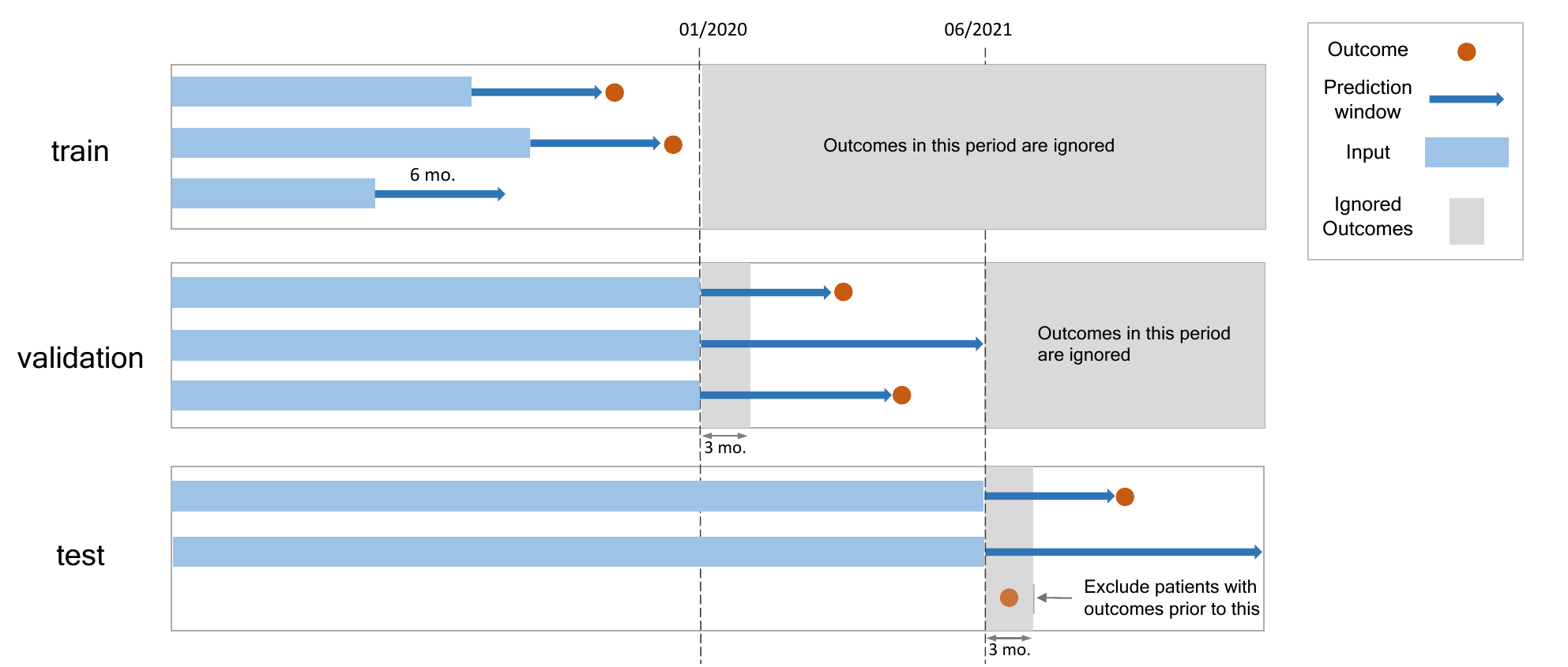}}
        \vspace{-1.5em}
	\caption{This figure illustrates the out-of-time fine-tuning and evaluation process. Training used data only up to 01/2020. The remaining three years were split for validation and testing: validation labels covered outcomes from 03/2020 to 06/2021, and testing labels covered outcomes from 09/2021 onward. Patients with outcomes before these dates in their respective sets were excluded.}
	\label{fig:out_of_time}
\end{figure}


\clearpage

\section{Cohort Selection} \label{appendix:cohort_selection}
\begin{figure}[H]
	\centering
	\makebox[\textwidth][c]{\includegraphics[scale=0.8]{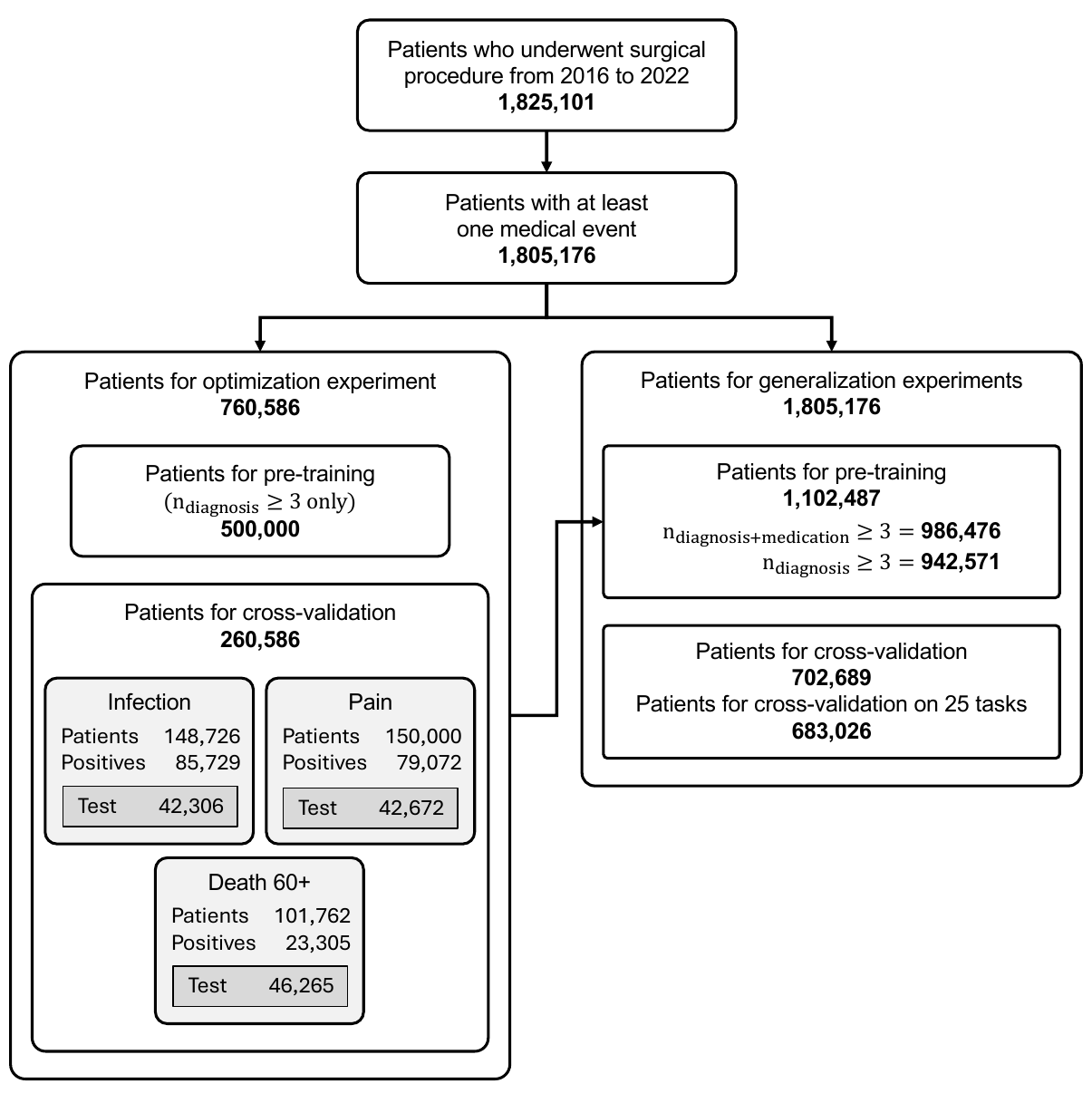}}
	\caption{Selection pipeline for model selection and final evaluation}
	\label{fig:cohort_overview}
\end{figure}

\section{Generalization Cohort Statistics}\label{appendix:generealization_stats}
\input{tables/cohort_stats_tables_v2}
\clearpage

\section{Age Distribution for Generalization Cohort}\label{appendix:age_distribution}
\begin{figure}[!h]
	\centering
	\includegraphics[width=.7\paperwidth]{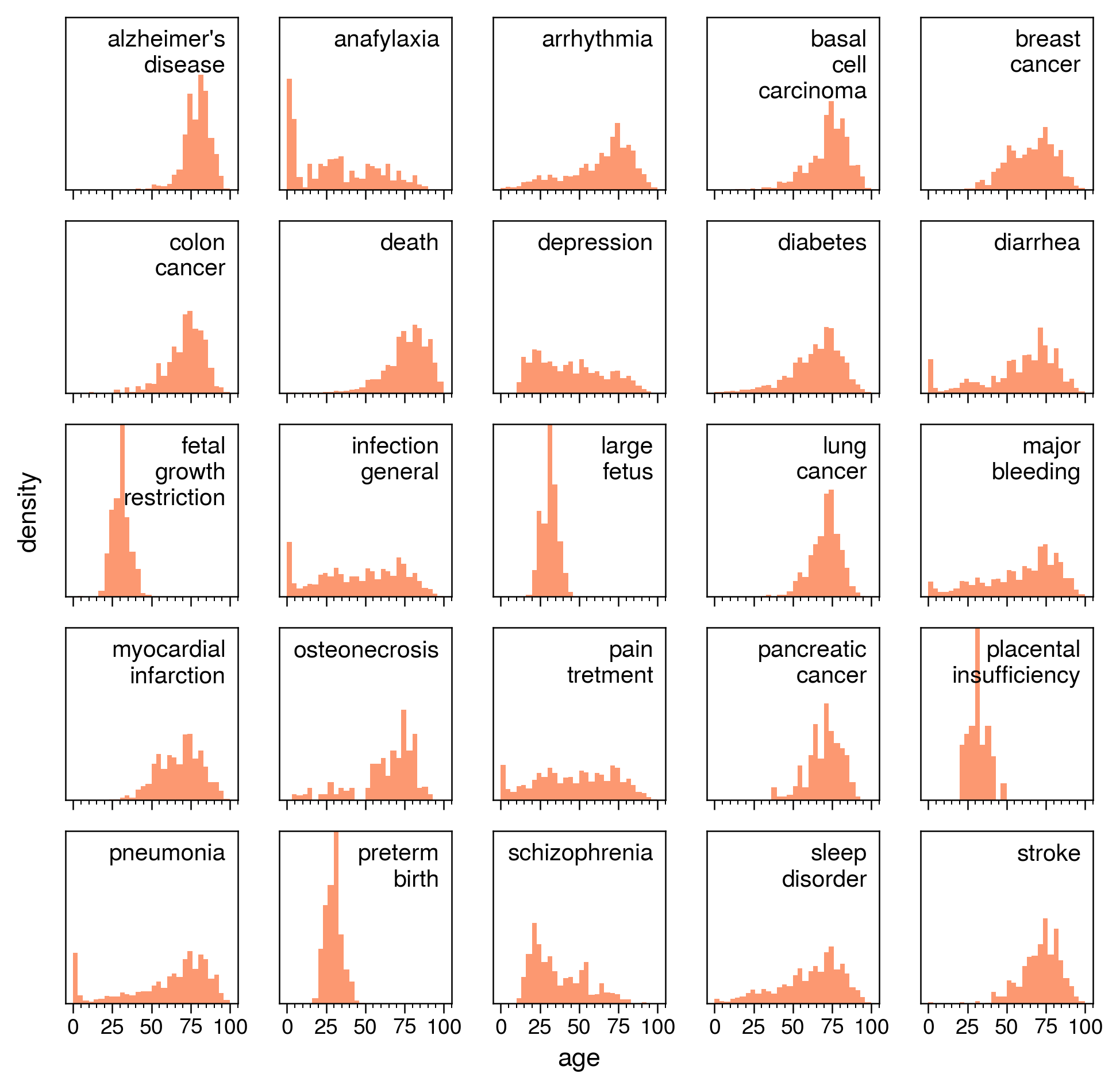}
	\caption{Age distribution for positive patients in the cross-validation set, before age group selection, for each of the 25 downstream tasks. }
\end{figure}
\clearpage

\section{Results of Generalization Experiments}\label{appendix:mr_and_heads_results}
\begin{figure}[H]
        \centering
	\makebox[\textwidth][c]{\includegraphics[scale=0.5]{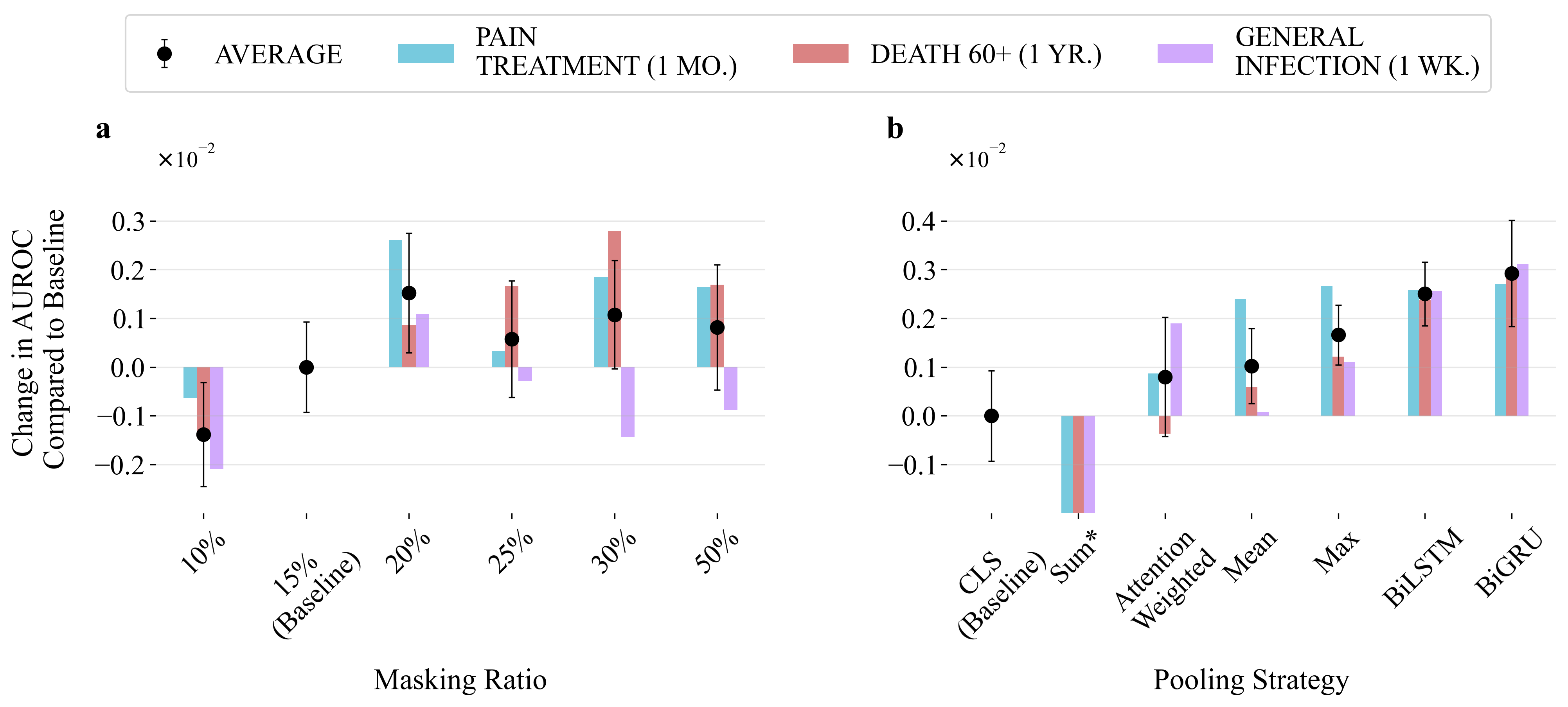}}
	\caption{Ablation studies analyzing the effects of different masking ratios (MR) and pooling strategies on predictive performance. In (\textbf{a}) we compare the change in AUROC across varying MRs using \behrtbest\;with 15\% MR and CLS (classification) token pooling as a baseline. While an MR of 20\% shows the highest average AUROC improvement, certain conditions such as death prediction benefit from a higher MR of 30\%. In (\textbf{b}) we evaluate various pooling strategies against \behrtbest\; with CLS pooling as a baseline.
    The Bidirectional Gated Recurrent Unit (BiGRU) pooling strategy emerges as superior based on the average AUROC improvement across conditions. \newline
    \small{* This change resulted in a change of average AUROC of $-1.5\cdot10^{-2}\,$, which is not displayed to maintain readability of other differences.}
    }
\end{figure}

\clearpage
\section{Results of Optimization Experiments}\label{appendix:initial_experiment}
\input{tables/initial_exp_results_v2}

\clearpage

\section{Definitions of Conditions}\label{appendix:disease_definitions}
\input{tables/disease_codes}

%% file: tables/cohort_stats_tables_v2.tex
\begin{table}\label{tab:generalization_cohort_stats}
\caption{Statistics for all 25 CV sets for the generalization experiment. sequence length is N events per patient sequence (post-censoring). Trajectory length* is in years; }
\begin{tabular}{
p{0.15\linewidth}
>{\centering\arraybackslash}p{0.15\linewidth}
>{\centering\arraybackslash}p{0.115\linewidth}
p{0.13\linewidth}
p{0.14\linewidth}
>{\centering\arraybackslash}p{0.09\linewidth}
>{\centering\arraybackslash}p{0.06\linewidth}
}
    \toprule
    Condition & Prediction Window & Age mean±std & Sequence\newline Length\newline Q2(Q1-Q3) & Trajectory Length \newline Q2(Q1-Q3)  & \multicolumn{2}{>{\centering\arraybackslash}p{0.09\linewidth}}{Patients\newline [thousands]/\newline Prevalence[\%]} \\
    \midrule
    Alzheimer's\newline Disease & 3 mo. & 79.1±7.1 & 27 (3-161) & 2.5 (1.3-3.9) & \dualcol{46.2}{3.7} \\
    Anaphylaxis & 24 hr. & 41.6±24.7 & 4 (1-28) & 2.2 (1.0-3.5) & \dualcol{338.1}{0.2} \\
    Arrhythmia & 6 mo. & 73.6±9.2 & 9 (2-85) & 2.4 (1.4-3.7) & \dualcol{70.9}{10.7} \\
    Basal Cell\newline Carcinoma & 1 yr. & 79.4±7.2 & 22 (3-147) & 3.0 (1.9-4.3) & \dualcol{41.2}{4.6} \\
    Breast \newline Cancer & 3 mo. & 74.4±9.6 & 11 (2-102) & 2.3 (1.2-3.8) & \dualcol{27.1}{4.5} \\
    Colon Cancer & 3 mo. & 73.7±9.3 & 14 (2-110) & 2.4 (1.3-3.9) & \dualcol{71.0}{1.3} \\
    Lung Cancer & 3 mo. & 73.7±9.2 & 14 (2-110) & 2.4 (1.3-3.9) & \dualcol{69.9}{2.1} \\
    Pancreatic Cancer & 3 mo. & 73.7±9.3 & 15 (3-117) & 2.5 (1.3-3.8) & \dualcol{75.5}{0.4} \\
    Death & 1 yr. & 73.7±9.1 & 11 (2-89) & 3.0 (2.0-4.3) & \dualcol{70.1}{22.5} \\
    Depression & 3 mo. & 22.5±4.3 & 2 (1-7) & 2.1 (1.1-3.4) & \dualcol{53.5}{2.7} \\
    Diabetes & 6 mo. & 41.9±24.7 & 3 (1-17) & 2.0 (1.1-3.1) & \dualcol{214.8}{5.0} \\
    Diarrhea & 3 d. & 41.5±24.7 & 4 (1-26) & 1.9 (0.9-3.1) & \dualcol{316.3}{1.3} \\
    Infection & 7 d. & 41.2±24.7 & 2 (1-7) & 1.4 (0.6-2.7) & \dualcol{281.9}{42.6} \\
    Major\newline Bleeding & 24 hr. & 41.7±24.7 & 4 (1-26) & 2.0 (0.9-3.4) & \dualcol{325.1}{4.7} \\
    Myocardial\newline Infarction & 14 d. & 73.7±9.3 & 16 (3-118) & 2.3 (1.1-3.7) & \dualcol{80.8}{2.7} \\
    Osteonecrosis & 6 mo. & 79.4±7.2 & 32 (4-176) & 3.0 (1.8-4.4) & \dualcol{51.4}{0.2} \\
    Pain\newline Treatment & 1 mo. & 40.8±24.5 & 2 (1-5) & 1.4 (0.6-2.5) & \dualcol{267.5}{42.0} \\
    Pneumonia & 14 d. & 41.7±24.8 & 4 (1-24) & 1.8 (0.8-3.3) & \dualcol{293.2}{5.5} \\
    Large Fetus & at pregnancy & 30.4±5.1 & 3 (2-7) & 0.5 (0.0-2.1) & \dualcol{22.2}{2.3} \\
    Placental\newline Insufficiency & at pregnancy & 30.4±5.0 & 3 (2-7) & 0.4 (0.0-2.1) & \dualcol{22.2}{80.6} \\
    Fetal Growth\newline Disorder & at pregnancy & 30.4±5.1 & 3 (2-7) & 0.4 (0.0-2.1) & \dualcol{22.2}{6.6} \\
    Preterm Birth & at pregnancy & 30.4±5.0 & 3 (2-7) & 0.4 (0.0-2.1) & \dualcol{22.2}{13.7} \\
    Schizophrenia & 3 mo. & 24.7±2.9 & 2 (1-7) & 1.6 (0.8-2.9) & \dualcol{25.5}{1.2} \\
    Sleep\newline Disorder & 3 mo. & 29.3±5.6 & 2 (1-11) & 1.8 (1.0-2.9) & \dualcol{67.9}{4.3} \\
    Stroke & 14 d. & 73.5±9.1 & 12 (2-102) & 1.9 (0.9-3.5) & \dualcol{60.4}{1.4} \\
    \bottomrule
    \multicolumn{5}{l}{\small{* From first encounter to censoring}}\\
\end{tabular}
\end{table}

%% file: tables/initial_exp_results_v2.tex
\renewcommand{\arraystretch}{1.2}
\begin{table}[H]
\caption{Test scores for initial experiments given as AUROC and AUPRC with mean and standard deviations obtained from the five models trained in CV for three prediction tasks pain treatment, death, and general infection. Our intermediate models are highlighted in bold. Adapted settings are shown with an asterisk.}
\label{tab:model_stats}
\centering
\small
\begin{tabular}{m{0.2\linewidth} p{0.105\linewidth} p{0.105\linewidth} p{0.105\linewidth} p{0.105\linewidth} p{0.105\linewidth} p{0.105\linewidth}}
\toprule
\multirow{2}{*}{Model} & \multicolumn{3}{c}{AUROC (\%)} & \multicolumn{3}{c}{AUPRC (\%)} \\
\cmidrule(lr){2-4} \cmidrule(lr){5-7}
& \multicolumn{1}{c}{Pain} & \multicolumn{1}{c}{Death} & \multicolumn{1}{c}{Infection} & \multicolumn{1}{c}{Pain} & \multicolumn{1}{c}{Death} & \multicolumn{1}{c}{Infection} \\
\midrule

\multicolumn{5}{p{0.5\linewidth}}{\textbf{Data representation}}  & & \\  
Med-BERT & 76.53$\pm$0.07 & 79.23$\pm$0.13 & 72.67$\pm$0.11 & 79.10$\pm$0.05 & 52.77$\pm$0.21 & 77.92$\pm$0.13 \\
\hspace{-.5em}*BEHRT & 77.01$\pm$0.13 & 84.54$\pm$0.15 & 73.87$\pm$0.07 & 79.51$\pm$0.17 & 60.27$\pm$0.35 & 78.84$\pm$0.06 \\
\hspace{-.5em}*+Medication & 77.74$\pm$0.08 & 84.62$\pm$0.06 & 74.37$\pm$0.10 & 80.28$\pm$0.07 & 60.91$\pm$0.23 & 79.50$\pm$0.09 \\
\hspace{-.5em}*+FV+Sex & 77.87$\pm$0.06 & 84.69$\pm$0.17 & 74.34$\pm$0.04 & 80.44$\pm$0.06 & 61.24$\pm$0.37 & 79.39$\pm$0.06 \\
-SEP & 77.80$\pm$0.02 & 84.21$\pm$0.17 & 74.37$\pm$0.07 & 80.30$\pm$0.05 & 61.13$\pm$0.39 & 79.34$\pm$0.06 \\
\hspace{-.5em}*Simplified Segments & 78.12$\pm$0.04 & 84.60$\pm$0.16 & 74.37$\pm$0.19 & 80.50$\pm$0.04 & 61.06$\pm$0.29 & 79.30$\pm$0.19 \\
\hspace{-.5em}*+t2v(timestamps) & 78.48$\pm$0.11 & 85.57$\pm$0.11 & 75.12$\pm$0.07 & 80.99$\pm$0.06 & 63.41$\pm$0.30 & 80.29$\pm$0.10 \\
$\mathrm{\textbf{BEHRT}}_{+\mathrm{\textbf{D}}}$ & & & & & & \\ \hline
\multicolumn{5}{p{0.5\linewidth}}{\textbf{Technical components}}  & & \\  
\hspace{-.5em}*+t2v(age) & 78.50$\pm$0.05 & 85.69$\pm$0.11 & 75.20$\pm$0.06 & 80.97$\pm$0.04 & 63.67$\pm$0.27 & 80.33$\pm$0.10 \\
\hspace{-.5em}*+RoPE+SwiGLU & 78.48$\pm$0.05 & 85.62$\pm$0.13 & 75.38$\pm$0.08 & 80.81$\pm$0.07 & 63.56$\pm$0.42 & 80.60$\pm$0.13 \\
Masking Ratio 10\% & 78.41$\pm$0.06 & 85.48$\pm$0.13 & 75.21$\pm$0.12 & 80.82$\pm$0.07 & 63.44$\pm$0.23 & 80.27$\pm$0.17 \\
\hspace{-.5em}*Masking Ratio 20\% & 78.74$\pm$0.12 & 85.77$\pm$0.08 & 75.57$\pm$0.13 & 81.08$\pm$0.09 & 63.89$\pm$0.23 & 80.62$\pm$0.10 \\
Masking Ratio 25\% & 78.64$\pm$0.06 & 85.94$\pm$0.15 & 75.24$\pm$0.11 & 80.96$\pm$0.08 & 64.22$\pm$0.19 & 80.44$\pm$0.09 \\
Masking Ratio 30\% & 78.56$\pm$0.06 & 85.74$\pm$0.19 & 75.41$\pm$0.11 & 80.89$\pm$0.03 & 63.94$\pm$0.42 & 80.42$\pm$0.14 \\
Masking Ratio 50\% & 78.23$\pm$0.02 & 85.21$\pm$0.20 & 74.72$\pm$0.12 & 80.56$\pm$0.02 & 63.28$\pm$0.32 & 79.78$\pm$0.12 \\
MR 20\% + PLOS & 78.66$\pm$0.13 & 85.68$\pm$0.08 & 75.69$\pm$0.10 & 80.94$\pm$0.15 & 63.44$\pm$0.45 & 80.65$\pm$0.13 \\
Max Pooling & 78.71$\pm$0.04 & 85.74$\pm$0.06 & 75.50$\pm$0.09 & 81.12$\pm$0.03 & 64.13$\pm$0.15 & 80.66$\pm$0.09 \\
Mean Pooling & 78.66$\pm$0.06 & 85.94$\pm$0.14 & 75.69$\pm$0.11 & 81.12$\pm$0.03 & 63.94$\pm$0.33 & 80.89$\pm$0.11 \\
Sum Pooling & 78.16$\pm$0.15 & 83.45$\pm$1.10 & 73.37$\pm$1.04 & 80.24$\pm$0.21 & 56.87$\pm$2.03 & 78.05$\pm$0.90 \\
Attention Weighted & 78.84$\pm$0.08 & 85.86$\pm$0.08 & 75.74$\pm$0.10 & 81.22$\pm$0.09 & 63.63$\pm$0.13 & 80.84$\pm$0.11 \\
BiLSTM & 78.74$\pm$0.06 & 85.74$\pm$0.08 & 75.54$\pm$0.06 & 81.18$\pm$0.09 & 64.22$\pm$0.20 & 80.62$\pm$0.09 \\
\hspace{-.5em}*BiGRU & 78.69$\pm$0.13 & 85.94$\pm$0.14 & 75.70$\pm$0.09 & 81.22$\pm$0.11 & 63.94$\pm$0.33 & 80.89$\pm$0.09 \\
\textbf{CORE-BEHRT} & & & & & & \\

\bottomrule
\end{tabular}
\end{table}

%% file: tables/disease_codes.tex
\begin{longtable}{p{0.25\linewidth}p{0.75\linewidth}}
\caption{Table of conditions and their corresponding codes as given in the SKS-browser, sorted alphabetically by condition.}\\
\toprule
Condition & Codes\\
\midrule
\endfirsthead
\multicolumn{2}{l}{\hspace{.5em}Table 7 (continued)}\\ \\
\toprule
Condition & Codes\\
\midrule
\endhead
    Alzheimer’s Disease & DG30, DF00 \\
    Anaphylaxis & DT782, DT788, DT789, DT882, DT886, DT881 \\
    Arrhythmia & DI47, DI49, DR00, MC01A, MC01B, Abnorm hjerterytme \\
    Basal Cell Carcinoma & DC44, anden hudkræft \textbf{exclude:} Unspecified Cancer\\
    Breast Cancer & DC50, brystkræft \textbf{exclude:} Unspecified Cancer \\
    Colon Cancer & DC18, kræft i tyktarmen \textbf{exclude:} Unspecified Cancer\\
    Depression & Depression, DF32, DF33 \\
    Diabetes & MA10BA02, MA10BD07, MA10BD08, MA10BD10, MA10BD11, MA10BD13, MA10BD19, MA10BD21, MA10BD24, MA10BH01, MA10BH02, MA10BH03, MA10BH04, MA10BH05, MA10A, MA10BB, MA10BK, MA10BJ, DE10, DE11, diabetes \\
    Diarrhea & DK529, DK521, MA07D \\
    Fetal Growth \newline Restriction & DO365 \\
    Fetal Macrosomia & DO366\\
    Lung Cancer & DC34, Kræft i bronkier eller lunge med metastaser, Kræft i lungens, kræft i lunge \textbf{exclude:} Unspecified Cancer \\
    Major bleeding & DD62, DI60, DI61, DI62, DI690, DI691, DJ942, DK250, DK252, DK254, DK256, DK260, DK262, DK264, DK276, DK270, DK272, DK274, DK276, DK280, DK282, DK284, DK286, DK290, DK298A, DK625, DK638C, DK920, DK921, DK922, DN02, DN93, DR04, DR31, DS064, DS065, DS066 \\
    Myocardial Infarction & DI21, Komplikationer i efterforløbet af akut myokardieinfarkt \\
    Osteonecrosis & DM87 \\
    Pain Treatment & MN02A, MN02B, MM02A\\
    Pancreatic Cancer & DC25, Kræft i caput pancreatis med metastaser, Kræft i cauda pancreatis, Kræft i corpus pancreatis \textbf{exclude:} Unspecified Cancer \\
    Placental Insufficiency & DO365A \\
    Pregnancy & DZ340, DZ348, DZ349 \\
    Preterm Birth & DO42, DO47, DO60 \\
    Schizophrenia & DF20 \\
    Sleep Disorder & DF51, DG47, MN05CF, MN05CH \\
    Stroke & DI64, Slagtilfælde uden oplysning om blødning eller infarkt \\
    Unspecified Cancer & Kræftsygdom UNS med metastaser, Kræft UNS overgribende flere lokalisationer med metastaser \\
    Infection &  DA, DB, DD738C, DD738D, DD762, DE060B, DG00-DG07, DG531, DG630, DG734, DG940, DI00-DI02, DI30, DI320, DI321, DI33, DI38, DI39, DI400-DI412, DI430, DI520, DI521, DK112, DK113, DK12, DK140, 
    DK230, DK770, DK858E, DK930, DM00, DM01, DM03, DM461-DM465, DM490-DM493, DM600, DM630, DM632, 
    DM650, DM651, DM680, DM710, DM711, DM726, DM730, DM731, DM86, DM900-DM902, DN61, 
    DN70-DN77, MJ01 
    DN080, DN109A-DN109C, DN12, DN136, DN151, DN160, DN288D, DN288E, DN288F, DN290, DN291, DN30, DN330, DN340, DN341, 
    DN342A, DN342B, DN390, DN412, DDN45, DN481, DN482, DN49, DN511, DN512 
    DA0, DK35, DDK37, DK57, DK65, DK67, MJ01, 
    DH60, DH610, DH620, DH621, DH623, DH65, DH66, DH670, DH671, DH68, DH70, DH730, DH750, DH830, DH940 
    DA46, DL00-DL08, DL303, DL738H, DL88 
    DA541, DB43, DD733, DE060A, DE236A, DE321, DG06, DG07, DH000A, DH050A, DH440A, DH600, DJ340A, DJ36, DJ383D, DJ387G, DJ390, DJ391, DJ398A, DJ851, DJ852, DJ853, DK113, DK122, DK130A, DK140A, DK209A, DK353A, DK353B, DK570, DK572, DK574, DK578, DK61, DK630, DK650, DK750, DK810A, DK858A, DL02, DL050, DL059, DM608A, DM868A, DM869A, DN151, DN340, DN412, DN450, DN482, DN492A, DN619A, DN619B, DN700A, DN700B, DN710A, DN730A, DN730B, DN732A, DN732B, DN733A, DN735A, DN738A, DN738C, DN751, DN764,  DN768A, 
    'infektiøse', 'infektion',\newline 
    \textbf{exclude}: DA06, DA07, DA33, DA630, DA65, DA66, DA67, DA68, DA69, DA881, DB079D, DB18, DB35, DB36, DB37, DB38, DB39, 
      DB4-DB7, DB80-DB83, DB85-DB92, DB94, DI300, DI521C, 
      DK112B, DM031A, DM036, DM863, DM864, DM865, DM866
      DN701, DN702, DN711, DN729B, DN729E, DN731, DN734, DN736, DN738B, DN75, DN750, DN758, DN758B, DN759, DN761A, DN761C, 
      DN761E, DN763A, DN765, DN766
      DN301, DN302, DN304, DN308, DN459B, DN481D
      DA06, DA07, DK573C, DK573D, DK574, DK574A, DK575, DK578, DK579, DK658A, DK658C, DK658E, DK658F, DK658G, DK658H 
      DH604, DH604A, DH605, DH605B, DH608, DH608A, DH652, DH653, DH654, DH654C, DH661, DH662, DH663, DH681, DH701, DH708
      DJ340E, DJ340F, DJ340G, DJ340H, 
      DL889, DL889C, DA541B, DB430, DB438, DB439, DK570B, DK570C, DK572B, DK572C, DK574A, DK650M, DK650N, DK650O, DK650P
\end{longtable}

%% file: full-paper-template.bbl
\begin{thebibliography}{53}
\providecommand{\natexlab}[1]{#1}
\providecommand{\url}[1]{\texttt{#1}}
\expandafter\ifx\csname urlstyle\endcsname\relax
  \providecommand{\doi}[1]{doi: #1}\else
  \providecommand{\doi}{doi: \begingroup \urlstyle{rm}\Url}\fi

\bibitem[AlSaad et~al.(2022)AlSaad, Malluhi, and Boughorbel]{alsaad2022predict_preterm_birth}
Rawan AlSaad, Qutaibah Malluhi, and Sabri Boughorbel.
\newblock Predictptb: an interpretable preterm birth prediction model using attention-based recurrent neural networks.
\newblock \emph{BioData Mining}, 15\penalty0 (1):\penalty0 6, 2022.

\bibitem[Arboix(2015)]{arboix2015cardiovascular}
Adri{\`a} Arboix.
\newblock Cardiovascular risk factors for acute stroke: Risk profiles in the different subtypes of ischemic stroke.
\newblock \emph{World Journal of Clinical Cases: WJCC}, 3\penalty0 (5):\penalty0 418, 2015.

\bibitem[Asan et~al.(2020)Asan, Bayrak, and Choudhury]{asan2020trust}
Onur Asan, Alparslan~Emrah Bayrak, and Avishek Choudhury.
\newblock Artificial intelligence and human trust in healthcare: Focus on clinicians.
\newblock \emph{J Med Internet Res.}, 2020.

\bibitem[Baevski et~al.(2020)Baevski, Zhou, Mohamed, and Auli]{baevski2020wav2vec2.0}
Alexei Baevski, Henry Zhou, Abdelrahman Mohamed, and Michael Auli.
\newblock wav2vec 2.0: A framework for self-supervised learning of speech representations, 2020.

\bibitem[Booth(1994)]{booth1994read}
Nick Booth.
\newblock What are the read codes?
\newblock \emph{Health libraries review}, 11\penalty0 (3):\penalty0 177--182, 1994.

\bibitem[Chen and Guestrin(2016)]{chen2016xgboost}
Tianqi Chen and Carlos Guestrin.
\newblock Xgboost: A scalable tree boosting system.
\newblock In \emph{Proceedings of the 22nd ACM SIGKDD International Conference on Knowledge Discovery and Data Mining}, KDD ’16. ACM, August 2016.
\newblock \doi{10.1145/2939672.2939785}.
\newblock URL \url{http://dx.doi.org/10.1145/2939672.2939785}.

\bibitem[Choi et~al.(2016{\natexlab{a}})Choi, Schuetz, Stewart, and Sun]{choi2016usingrn}
E.~Choi, Andy Schuetz, Walter~F. Stewart, and Jimeng Sun.
\newblock Using recurrent neural network models for early detection of heart failure onset.
\newblock \emph{Journal of the American Medical Informatics Association : JAMIA}, 24:\penalty0 361 -- 370, 2016{\natexlab{a}}.
\newblock URL \url{https://api.semanticscholar.org/CorpusID:3763233}.

\bibitem[Choi et~al.(2015)Choi, Bahadori, and Sun]{choi2015doctorai}
Edward Choi, Mohammad~Taha Bahadori, and Jimeng Sun.
\newblock Doctor {AI:} predicting clinical events via recurrent neural networks.
\newblock \emph{CoRR}, abs/1511.05942, 2015.

\bibitem[Choi et~al.(2016{\natexlab{b}})Choi, Bahadori, Schuetz, Stewart, and Sun]{choi2016retain}
Edward Choi, Mohammad~Taha Bahadori, Andy Schuetz, Walter~F. Stewart, and Jimeng Sun.
\newblock {RETAIN:} interpretable predictive model in healthcare using reverse time attention mechanism.
\newblock \emph{CoRR}, abs/1608.05745, 2016{\natexlab{b}}.

\bibitem[Chowdhery et~al.(2022)Chowdhery, Narang, Devlin, Bosma, Mishra, Roberts, Barham, Chung, Sutton, Gehrmann, Schuh, Shi, Tsvyashchenko, Maynez, Rao, Barnes, Tay, Shazeer, Prabhakaran, Reif, Du, Hutchinson, Pope, Bradbury, Austin, Isard, Gur-Ari, Yin, Duke, Levskaya, Ghemawat, Dev, Michalewski, Garcia, Misra, Robinson, Fedus, Zhou, Ippolito, Luan, Lim, Zoph, Spiridonov, Sepassi, Dohan, Agrawal, Omernick, Dai, Pillai, Pellat, Lewkowycz, Moreira, Child, Polozov, Lee, Zhou, Wang, Saeta, Diaz, Firat, Catasta, Wei, Meier-Hellstern, Eck, Dean, Petrov, and Fiedel]{chowdhery2022palm}
Aakanksha Chowdhery, Sharan Narang, Jacob Devlin, Maarten Bosma, Gaurav Mishra, Adam Roberts, Paul Barham, Hyung~Won Chung, Charles Sutton, Sebastian Gehrmann, Parker Schuh, Kensen Shi, Sasha Tsvyashchenko, Joshua Maynez, Abhishek Rao, Parker Barnes, Yi~Tay, Noam Shazeer, Vinodkumar Prabhakaran, Emily Reif, Nan Du, Ben Hutchinson, Reiner Pope, James Bradbury, Jacob Austin, Michael Isard, Guy Gur-Ari, Pengcheng Yin, Toju Duke, Anselm Levskaya, Sanjay Ghemawat, Sunipa Dev, Henryk Michalewski, Xavier Garcia, Vedant Misra, Kevin Robinson, Liam Fedus, Denny Zhou, Daphne Ippolito, David Luan, Hyeontaek Lim, Barret Zoph, Alexander Spiridonov, Ryan Sepassi, David Dohan, Shivani Agrawal, Mark Omernick, Andrew~M. Dai, Thanumalayan~Sankaranarayana Pillai, Marie Pellat, Aitor Lewkowycz, Erica Moreira, Rewon Child, Oleksandr Polozov, Katherine Lee, Zongwei Zhou, Xuezhi Wang, Brennan Saeta, Mark Diaz, Orhan Firat, Michele Catasta, Jason Wei, Kathy Meier-Hellstern, Douglas Eck, Jeff Dean, Slav Petrov, and Noah Fiedel.
\newblock Palm: Scaling language modeling with pathways, 2022.

\bibitem[Dauphin et~al.(2017)Dauphin, Fan, Auli, and Grangier]{dauphin2017glu}
Yann~N. Dauphin, Angela Fan, Michael Auli, and David Grangier.
\newblock Language modeling with gated convolutional networks, 2017.

\bibitem[Dettmers et~al.(2023)Dettmers, Pagnoni, Holtzman, and Zettlemoyer]{dettmers2023qlora}
Tim Dettmers, Artidoro Pagnoni, Ari Holtzman, and Luke Zettlemoyer.
\newblock Qlora: Efficient finetuning of quantized llms, 2023.

\bibitem[Devlin et~al.(2018)Devlin, Chang, Lee, and Toutanova]{devlin2018bert}
Jacob Devlin, Ming-Wei Chang, Kenton Lee, and Kristina Toutanova.
\newblock Bert: Pre-training of deep bidirectional transformers for language understanding.
\newblock \emph{arXiv preprint arXiv:1810.04805}, 2018.

\bibitem[Dosovitskiy et~al.(2021)Dosovitskiy, Beyer, Kolesnikov, Weissenborn, Zhai, Unterthiner, Dehghani, Minderer, Heigold, Gelly, Uszkoreit, and Houlsby]{dosovitskiy2021vit}
Alexey Dosovitskiy, Lucas Beyer, Alexander Kolesnikov, Dirk Weissenborn, Xiaohua Zhai, Thomas Unterthiner, Mostafa Dehghani, Matthias Minderer, Georg Heigold, Sylvain Gelly, Jakob Uszkoreit, and Neil Houlsby.
\newblock An image is worth 16x16 words: Transformers for image recognition at scale, 2021.

\bibitem[He et~al.(2021)He, Chen, Xie, Li, Dollár, and Girshick]{he2021mae}
Kaiming He, Xinlei Chen, Saining Xie, Yanghao Li, Piotr Dollár, and Ross Girshick.
\newblock Masked autoencoders are scalable vision learners, 2021.

\bibitem[Henriksen et~al.(2017)Henriksen, Nordgaard, and Jansson]{henriksen2017genetics_schizophrenia}
Mads~G Henriksen, Julie Nordgaard, and Lennart~B Jansson.
\newblock Genetics of schizophrenia: overview of methods, findings and limitations.
\newblock \emph{Frontiers in human neuroscience}, 11:\penalty0 322, 2017.

\bibitem[Horsky et~al.(2017)Horsky, Drucker, and Ramelson]{horsky2017accuracy_icd10}
Jan Horsky, Elizabeth~A Drucker, and Harley~Z Ramelson.
\newblock Accuracy and completeness of clinical coding using icd-10 for ambulatory visits.
\newblock In \emph{AMIA annual symposium proceedings}, volume 2017, page 912. American Medical Informatics Association, 2017.

\bibitem[Johnson et~al.(2016)Johnson, A., Pollard, T., Shen, and et~al.]{johnson2016mimic3}
Johnson, A., Pollard, T., Shen, and L.~et~al.
\newblock Mimic-iii, a freely accessible critical care database.
\newblock \emph{Scientific Data}, 2016.

\bibitem[Kazemi et~al.(2019)Kazemi, Goel, Eghbali, Ramanan, Sahota, Thakur, Wu, Smyth, Poupart, and Brubaker]{kazemi2019time2vec}
Seyed~Mehran Kazemi, Rishab Goel, Sepehr Eghbali, Janahan Ramanan, Jaspreet Sahota, Sanjay Thakur, Stella Wu, Cathal Smyth, Pascal Poupart, and Marcus Brubaker.
\newblock Time2vec: Learning a vector representation of time, 2019.

\bibitem[Khalilia et~al.(2011)Khalilia, Chakraborty, and Popescu]{khalilia2011predicting_rf}
Mohammed Khalilia, Sounak Chakraborty, and Mihail Popescu.
\newblock Predicting disease risks from highly imbalanced data using random forest.
\newblock \emph{BMC medical informatics and decision making}, 11:\penalty0 1--13, 2011.

\bibitem[Lauritzen et~al.(2023)Lauritzen, von Euler-Chelpin, Lynge, Vejborg, Nielsen, Karssemeijer, and Lillholm]{lauritzen2023assessing}
Andreas~D Lauritzen, My~C von Euler-Chelpin, Elsebeth Lynge, Ilse Vejborg, Mads Nielsen, Nico Karssemeijer, and Martin Lillholm.
\newblock Assessing breast cancer risk by combining ai for lesion detection and mammographic texture.
\newblock \emph{Radiology}, 308\penalty0 (2):\penalty0 e230227, 2023.

\bibitem[Li et~al.(2020)Li, Rao, Solares, Hassaine, Ramakrishnan, Canoy, Zhu, Rahimi, and Salimi-Khorshidi]{li2020behrt}
Yikuan Li, Shishir Rao, Jos{\'e} Roberto~Ayala Solares, Abdelaali Hassaine, Rema Ramakrishnan, Dexter Canoy, Yajie Zhu, Kazem Rahimi, and Gholamreza Salimi-Khorshidi.
\newblock Behrt: transformer for electronic health records.
\newblock \emph{Scientific reports}, 10\penalty0 (1):\penalty0 1--12, 2020.

\bibitem[Li et~al.(2022)Li, Mamouei, Salimi-Khorshidi, Rao, Hassaine, Canoy, Lukasiewicz, and Rahimi]{li2022hibehrt}
Yikuan Li, Mohammad Mamouei, Gholamreza Salimi-Khorshidi, Shishir Rao, Abdelaali Hassaine, Dexter Canoy, Thomas Lukasiewicz, and Kazem Rahimi.
\newblock Hi-behrt: Hierarchical transformer-based model for accurate prediction of clinical events using multimodal longitudinal electronic health records.
\newblock \emph{IEEE Journal of Biomedical and Health Informatics}, 2022.

\bibitem[Liu et~al.(2019)Liu, Ott, Goyal, Du, Joshi, Chen, Levy, Lewis, Zettlemoyer, and Stoyanov]{liu2019roberta}
Yinhan Liu, Myle Ott, Naman Goyal, Jingfei Du, Mandar Joshi, Danqi Chen, Omer Levy, Mike Lewis, Luke Zettlemoyer, and Veselin Stoyanov.
\newblock Roberta: A robustly optimized bert pretraining approach, 2019.

\bibitem[Mahajan et~al.(2018)Mahajan, Girshick, Ramanathan, He, Paluri, Li, Bharambe, and van~der Maaten]{mahajan2018sqrt}
Dhruv Mahajan, Ross~B. Girshick, Vignesh Ramanathan, Kaiming He, Manohar Paluri, Yixuan Li, Ashwin Bharambe, and Laurens van~der Maaten.
\newblock Exploring the limits of weakly supervised pretraining.
\newblock \emph{CoRR}, abs/1805.00932, 2018.
\newblock URL \url{http://arxiv.org/abs/1805.00932}.

\bibitem[Mikolov et~al.(2013)Mikolov, Sutskever, Chen, Corrado, and Dean]{mikolov2013nlpsqrt}
Tom{\'{a}}s Mikolov, Ilya Sutskever, Kai Chen, Greg Corrado, and Jeffrey Dean.
\newblock Distributed representations of words and phrases and their compositionality.
\newblock \emph{CoRR}, abs/1310.4546, 2013.
\newblock URL \url{http://arxiv.org/abs/1310.4546}.

\bibitem[MJ et~al.(1997)MJ, TE, DM, BH, LA, DE, CM, TJ, and WN]{fine1997logisticregression}
Fine MJ, Auble TE, Yealy DM, Hanusa BH, Weissfeld LA, Singer DE, Coley CM, Marrie TJ, and Kapoor WN.
\newblock A prediction rule to identify low-risk patients with community-acquired pneumonia.
\newblock \emph{N Engl J Med.}, 1997.

\bibitem[Montagnon et~al.(2020)Montagnon, Cerny, Cadrin-Ch{\^e}nevert, Hamilton, Derennes, Ilinca, Vandenbroucke-Menu, Turcotte, Kadoury, and Tang]{montagnon2020deep_radiology}
Emmanuel Montagnon, Milena Cerny, Alexandre Cadrin-Ch{\^e}nevert, Vincent Hamilton, Thomas Derennes, Andr{\'e} Ilinca, Franck Vandenbroucke-Menu, Simon Turcotte, Samuel Kadoury, and An~Tang.
\newblock Deep learning workflow in radiology: a primer.
\newblock \emph{Insights into imaging}, 11:\penalty0 1--15, 2020.

\bibitem[Nguyen et~al.(2016)Nguyen, Tran, Wickramasinghe, and Venkatesh]{nguyen2016deepr}
Phuoc Nguyen, Truyen Tran, Nilmini Wickramasinghe, and Svetha Venkatesh.
\newblock $\backslash$ mathtt $\{$Deepr$\}$ : a convolutional net for medical records.
\newblock \emph{IEEE journal of biomedical and health informatics}, 21\penalty0 (1):\penalty0 22--30, 2016.

\bibitem[Nourazari et~al.(2021)Nourazari, Davis, Granovsky, Austin, Straff, Joseph, and Sanchez]{nourazari2021decreased}
Sara Nourazari, Samuel~R Davis, Rachel Granovsky, Randolph Austin, Dean~J Straff, Joshua~W Joseph, and Leon~D Sanchez.
\newblock Decreased hospital admissions through emergency departments during the covid-19 pandemic.
\newblock \emph{The American journal of emergency medicine}, 42:\penalty0 203--210, 2021.

\bibitem[Pang et~al.(2021)Pang, Jiang, Kalluri, Spotnitz, Chen, Perotte, and Natarajan]{pang2021cehr}
Chao Pang, Xinzhuo Jiang, Krishna~S Kalluri, Matthew Spotnitz, RuiJun Chen, Adler Perotte, and Karthik Natarajan.
\newblock Cehr-bert: Incorporating temporal information from structured ehr data to improve prediction tasks.
\newblock In \emph{Machine Learning for Health}, pages 239--260. PMLR, 2021.

\bibitem[Prakash et~al.(2021)Prakash, Chilukuri, Ranade, and Viswanathan]{prakash2021rarebert}
P.~Prakash, S.~Chilukuri, N.~Ranade, and S.~Viswanathan.
\newblock Rarebert: Transformer architecture for rare disease patient identification using administrative claims.
\newblock \emph{Proceedings of the AAAI Conference on Artificial Intelligence}, 2021.

\bibitem[Ramachandran et~al.(2017)Ramachandran, Zoph, and Le]{ramachandran2017swish}
Prajit Ramachandran, Barret Zoph, and Quoc~V. Le.
\newblock Searching for activation functions, 2017.

\bibitem[Rao et~al.(2022)Rao, Mamouei, Salimi-Khorshidi, Li, Ramakrishnan, Hassaine, Canoy, and Rahimi]{rao2022targetedbehrt}
Shishir Rao, Mohammad Mamouei, Gholamreza Salimi-Khorshidi, Yikuan Li, Rema Ramakrishnan, Abdelaali Hassaine, Dexter Canoy, and Kazem Rahimi.
\newblock Targeted-behrt: Deep learning for observational causal inference on longitudinal electronic health records.
\newblock \emph{IEEE Transactions on Neural Networks and Learning Systems}, 2022.

\bibitem[Rasmy et~al.(2021)Rasmy, Xiang, Xie, Tao, and Zhi]{rasmy2021medbert}
Laila Rasmy, Yang Xiang, Ziqian Xie, Cui Tao, and Degui Zhi.
\newblock Med-bert: pretrained contextualized embeddings on large-scale structured electronic health records for disease prediction.
\newblock \emph{NPJ digital medicine}, 4\penalty0 (1):\penalty0 86, 2021.

\bibitem[Schuster and Paliwal(1997)]{schuster1997bidirectional}
Mike Schuster and Kuldip~K Paliwal.
\newblock Bidirectional recurrent neural networks.
\newblock \emph{IEEE transactions on Signal Processing}, 45\penalty0 (11):\penalty0 2673--2681, 1997.

\bibitem[Shang et~al.(2019)Shang, Ma, Xiao, and Sun]{shang2019gbert}
Junyuan Shang, Tengfei Ma, Cao Xiao, and Jimeng Sun.
\newblock Pre-training of graph augmented transformers for medication recommendation.
\newblock \emph{arXiv preprint arXiv:1906.00346}, 2019.

\bibitem[Shazeer(2020)]{shazeer2020glu}
Noam Shazeer.
\newblock Glu variants improve transformer, 2020.

\bibitem[Slawomirski et~al.(2023)Slawomirski, Lindner, de~Bienassis, Haywood, Hashiguchi, Steentjes, and Oderkirk]{slawomirski2023progress}
Luke Slawomirski, Luca Lindner, Katherine de~Bienassis, Philip Haywood, Tiago Cravo~Oliveira Hashiguchi, Melanie Steentjes, and Jillian Oderkirk.
\newblock Progress on implementing and using electronic health record systems: Developments in oecd countries as of 2021.
\newblock \emph{OECD Health Working Papers}, 2023.

\bibitem[Stilo and Murray(2019)]{stilo2019non_genetic_schizophrenia}
Simona~A Stilo and Robin~M Murray.
\newblock Non-genetic factors in schizophrenia.
\newblock \emph{Current psychiatry reports}, 21:\penalty0 1--10, 2019.

\bibitem[Su et~al.(2023)Su, Lu, Pan, Murtadha, Wen, and Liu]{su2023roformer}
Jianlin Su, Yu~Lu, Shengfeng Pan, Ahmed Murtadha, Bo~Wen, and Yunfeng Liu.
\newblock Roformer: Enhanced transformer with rotary position embedding, 2023.

\bibitem[Teoh(2018)]{teoh2018towards_stroke_prediction}
Douglas Teoh.
\newblock Towards stroke prediction using electronic health records.
\newblock \emph{BMC medical informatics and decision making}, 18:\penalty0 1--11, 2018.

\bibitem[Tong et~al.(2022)Tong, Song, Wang, and Wang]{tong2022videomae}
Zhan Tong, Yibing Song, Jue Wang, and Limin Wang.
\newblock Videomae: Masked autoencoders are data-efficient learners for self-supervised video pre-training, 2022.

\bibitem[Touvron et~al.(2023)Touvron, Lavril, Izacard, Martinet, Lachaux, Lacroix, Rozière, Goyal, Hambro, Azhar, Rodriguez, Joulin, Grave, and Lample]{touvron2023llama}
Hugo Touvron, Thibaut Lavril, Gautier Izacard, Xavier Martinet, Marie-Anne Lachaux, Timothée Lacroix, Baptiste Rozière, Naman Goyal, Eric Hambro, Faisal Azhar, Aurelien Rodriguez, Armand Joulin, Edouard Grave, and Guillaume Lample.
\newblock Llama: Open and efficient foundation language models, 2023.

\bibitem[V et~al.(2019)V, S, A, K, O, S, S, M, D, CA, RT, R, R, JP, ICK, H, and AD]{kuan2019caliber}
Kuan V, Denaxas S, Gonzalez-Izquierdo A, Direk K, Bhatti O, Husain S, Sutaria S, Hingorani M, Nitsch D, Parisinos CA, Lumbers RT, Mathur R, Sofat R, Casas JP, Wong ICK, Hemingway H, and Hingorani AD.
\newblock A chronological map of 308 physical and mental health conditions from 4 million individuals in the english national health service.
\newblock \emph{Lancet Digit Health}, 2019.

\bibitem[Vaswani et~al.(2017)Vaswani, Shazeer, Parmar, Uszkoreit, Jones, Gomez, Kaiser, and Polosukhin]{vaswani2017attention}
Ashish Vaswani, Noam Shazeer, Niki Parmar, Jakob Uszkoreit, Llion Jones, Aidan~N Gomez, {\L}ukasz Kaiser, and Illia Polosukhin.
\newblock Attention is all you need.
\newblock \emph{Advances in neural information processing systems}, 30, 2017.

\bibitem[Wei et~al.(2016)Wei, Teixeira, Mo, Cronin, Warner, and Denny]{wei2016combining_ehr}
Wei-Qi Wei, Pedro~L Teixeira, Huan Mo, Robert~M Cronin, Jeremy~L Warner, and Joshua~C Denny.
\newblock Combining billing codes, clinical notes, and medications from electronic health records provides superior phenotyping performance.
\newblock \emph{Journal of the American Medical Informatics Association}, 23\penalty0 (e1):\penalty0 e20--e27, 2016.

\bibitem[Wettig et~al.(2023)Wettig, Gao, Zhong, and Chen]{wettig2023mask}
Alexander Wettig, Tianyu Gao, Zexuan Zhong, and Danqi Chen.
\newblock Should you mask 15

\bibitem[Wu et~al.(2017)Wu, Burnside, Cox, Fan, Yuan, Yin, Peissig, Cobian, Page, and Craven]{wu2017breast_cancer}
Yirong Wu, Elizabeth~S Burnside, Jennifer Cox, Jun Fan, Ming Yuan, Jie Yin, Peggy Peissig, Alexander Cobian, David Page, and Mark Craven.
\newblock Breast cancer risk prediction using electronic health records.
\newblock In \emph{2017 IEEE International Conference on Healthcare Informatics (ICHI)}, pages 224--228. IEEE, 2017.

\bibitem[Y et~al.(2021)Y, W, MK, and CW.]{meng2021brltm}
Meng Y, Speier W, Ong MK, and Arnold CW.
\newblock Bidirectional representation learning from transformers using multimodal electronic health record data to predict depression.
\newblock \emph{IEEE Journal of Biomedical and Health Informatics}, 2021.

\bibitem[Yang et~al.(2019)Yang, Dai, Yang, Carbonell, Salakhutdinov, and Le]{yang2019xlnet}
Zhilin Yang, Zihang Dai, Yiming Yang, Jaime Carbonell, Russ~R Salakhutdinov, and Quoc~V Le.
\newblock Xlnet: Generalized autoregressive pretraining for language understanding.
\newblock \emph{Advances in neural information processing systems}, 32, 2019.

\bibitem[Yasaka and Abe(2018)]{yasaka2018deep_radiology}
Koichiro Yasaka and Osamu Abe.
\newblock Deep learning and artificial intelligence in radiology: Current applications and future directions.
\newblock \emph{PLoS medicine}, 15\penalty0 (11):\penalty0 e1002707, 2018.

\bibitem[Zhang et~al.(2022)Zhang, Chen, and Bui]{zhang2022adadiag}
Tianran Zhang, Muhao Chen, and Alex~A.T. Bui.
\newblock Adadiag: Adversarial domain adaptation of diagnostic prediction with clinical event sequences,.
\newblock \emph{Journal of Biomedical Informatics}, 2022.

\end{thebibliography}
